\documentclass[conference]{IEEEtran}
\usepackage{times}

\usepackage{multicol}
\usepackage[bookmarks=true]{hyperref}
\usepackage{color}
\usepackage[normalem]{ulem}
\usepackage[numbers, sort&compress]{natbib}
\usepackage{graphicx}
\usepackage{subcaption} 

\usepackage{graphicx}
\usepackage{subcaption}
\usepackage{mathtools}
\usepackage{multirow}
\usepackage{booktabs}

\newcommand{\shm}{\textcolor{magenta}}
\newcommand{\shb}{\textcolor{brown}}
\newcommand{\bhr}{\textcolor{red}}

\usepackage{booktabs}
\usepackage{graphicx}
\usepackage{xcolor}
\usepackage{hyperref}
\usepackage{etoolbox}
\usepackage{capt-of}
\IEEEoverridecommandlockouts
\usepackage[compatibility=false]{caption}
\usepackage{amsmath}
\usepackage{amssymb}
\newcounter{boldpara}
\newcommand{\boldparagraph}[1]{
    \refstepcounter{boldpara}
    \vspace{0.1em}\noindent{\bf #1}
}
\usepackage{dblfloatfix}
\let\labelindent\relax 
\usepackage[inline]{enumitem}

\usepackage[table]{xcolor}
\definecolor{IncBlue}{RGB}{31,119,180}
\definecolor{DecRed}{RGB}{214,39,40}
\definecolor{OursRow}{gray}{0.93} 

\usepackage[normalem]{ulem}

\newcommand{\jb}[1]{\textcolor{orange}{\textbf{jb-}#1}}

\IEEEaftertitletext{%
  \begin{center}
    \includegraphics[width=\textwidth]{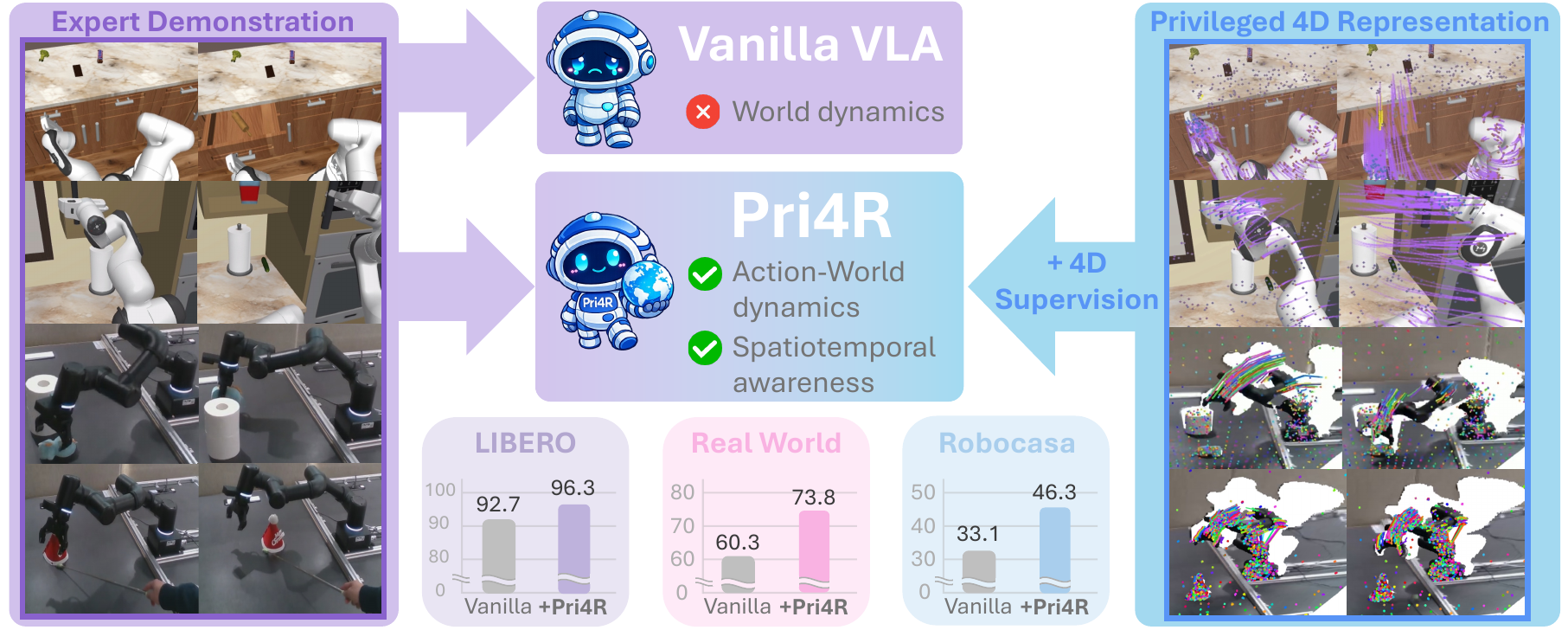}
\captionof{figure}{\textbf{Pri4R} equips Vision--Language--Action (VLA) models with an implicit awareness of action--world dynamics via \emph{privileged} 4D geometric supervision. Unlike standard VLAs (top) trained only by action imitation (left), Pri4R extracts 3D point tracks from demonstrations (right) and adds an auxiliary head to predict future point trajectories alongside actions. This training signal encourages the policy to model how scene geometry evolves under interaction, improving robustness and task success (bottom), while preserving the original test-time interface with zero inference overhead.}
    \label{fig:teaser}
  \end{center}
}

\begin{document}

\title{Pri4R: Learning World Dynamics \\ for Vision-Language-Action Models \\ with Privileged 4D Representation}


\author{
  Jisoo Kim$^{\ast1,2}$, Jungbin Cho$^{\ast3,5}$, Sanghyeok Chu$^{2,4}$, Ananya Bal$^{5}$,
  Jinhyung Kim$^2$, Gunhee Lee$^2$, Sihaeng Lee$^2$, \\ Seung Hwan Kim$^2$
  Bohyung Han$^{\dagger4}$, Hyunmin Lee$^{\dagger2}$, Laszlo A. Jeni$^{\dagger5}$, Seungryong Kim$^{\dagger1}$ \\
  $^1$KAIST AI
  $^2$LG AI Research,
  $^3$Yonsei University,
  $^4$Seoul National University,
  $^5$Carnegie Mellon University \\
  \url{https://jiiiisoo.github.io/Pri4R/}}



%

\maketitle

\begin{abstract}

Humans learn not only how their bodies move, but also how the surrounding world responds to their actions. 
In contrast, while recent Vision-Language-Action (VLA) models exhibit impressive semantic understanding, they often fail to capture the spatiotemporal dynamics governing physical interaction. 
In this paper, we introduce Pri4R, a simple yet effective approach that endows VLA models with an implicit understanding of world dynamics by leveraging privileged 4D information during training.
Specifically, Pri4R augments VLAs with a lightweight point track head that predicts 3D point tracks. 
By injecting VLA features into this head to jointly predict future 3D trajectories, the model learns to incorporate the evolving scene geometry within its shared representation space. 
This allows the action prediction components to leverage a more physically-aware context for precise control. 
Due to its architectural simplicity, Pri4R is seamlessly compatible with dominant VLA design patterns through minimal changes. 
During inference, we run the model using the original VLA architecture unchanged; Pri4R adds no extra inputs, outputs, or computational overhead during inference.
Across simulation and real-world evaluations, Pri4R significantly improves performance on challenging manipulation tasks, including a +10\% gain on LIBERO-Long and a +40\% gain on RoboCasa. 
We further show that 3D point track prediction is an effective supervision target for learning action-world dynamics, while also validating our design choices through extensive ablations. Code and checkpoints will be released.

\end{abstract}

\IEEEpeerreviewmaketitle

\section{Introduction}
 
Generalizable robot policies require not only semantic understanding of the surrounding environment, but also knowledge of the dynamics governing action–world interactions. 
Recent advances in Vision–Language Models (VLMs)~\cite{llava, flamingo}, together with the emergence of large-scale robot datasets~\cite{bridge, droid, openx}, have enabled Vision–Language–Action (VLA) models~\cite{kim2024openvla, intelligence2025pi_, pi_0} to leverage pretrained VLMs for robotic control. 
A key design principle of these approaches is to minimize architectural changes, introducing only a lightweight action head to transfer the high-level semantic capabilities of VLMs to robot manipulation. 

However, this adaptation uses a sole learning signal: simple action labels, which primarily encourage \textit{imitation of demonstrated motions}, with no knowledge about \textit{the dynamics of the world}. 
\textbf{Put simply, action labels specify \emph{how to move}, but not \emph{what will happen}.} As a result, the learned policies often produce semantically plausible actions that lack an understanding of underlying world dynamics (e.g., attempting to grasp a handle without accounting for the door’s kinematic constraints), leading to brittle or inaccurate object interactions and ultimately, task failure.

In contrast, humans reason about actions through an internal understanding of geometry and dynamics, anticipating \textit{how their embodiment and surrounding objects will move, deform, or make contact in response to actions}~\cite{internal_model_for_sensorimotor_integration, forward_models, the_role_of_internal_models, foward_modeling_allows_feedback, sensory_input_and_control_of_grip}. 
Some efforts sought to incorporate such predictive capabilities by introducing additional models that generate future images or states~\cite{hu2024video, bharadhwaj2024gen2act, du2023learning, wu2023unleashing}. 
More recent approaches instead train a uniform policy to predict actions together with additional signals, including high-level abstractions (e.g., language~\cite{zhou2025chatvla, shi2025hi}, feature embeddings~\cite{zhangdreamvla, bu2025agibot}, object-centric representations~\cite{zhao2025cot, intelligence2025pi_}), and even dense observations (e.g., images~\cite{zhao2025cot, tian2024predictive} and depth maps~\cite{zhangdreamvla}).

Although these signals provide extra knowledge for action prediction, they are not directly aligned with the \textit{spatiotemporal metric space} in which action--world interactions unfold, yielding only indirect supervision where the model must still learn to utilize these cues for precise control. 
This mismatch suggests that truly modeling world dynamics requires a representation that is both metric and temporally grounded, namely \textit{how 3D geometry evolves over time (4D)}, rather than predicting static abstractions or goal observations. Furthermore, the aforementioned methods require additional computation during inference, along with complex design choices and hyperparameters.

In this paper, we propose \textbf{Pri4R}, a simple yet effective framework that equips VLA models with an implicit knowledge of world dynamics, enabling more accurate and robust behavior. 
The key idea is to utilize 4D geometry as a privileged supervision signal during training to revise the VLM's latent representations. 
By injecting the VLM's internal embeddings into a dedicated point tracking head, we guide the backbone to encode the causal relationship between actions and the resulting geometric evolution of the scene. 
This shared, dynamics-aware representation, then directly empowers the action prediction head to reason with a deeper physical context, without introducing additional inputs or architectural changes at inference time.
Figure~\ref{fig:teaser} illustrates the concept of Pri4R.

Specifically, we first precompute 3D point tracks for the demonstration trajectories used for training. 
We then augment the VLA with a lightweight point tracking head that predicts future 3D trajectories, conditioned on the model's internal embeddings. 
This design ensures that the gradients from the point tracking task inform and enrich the shared feature space used for action prediction. 
Pri4R is broadly applicable and easy to integrate into existing VLA frameworks with minimal architectural changes, for both backbone-centric VLAs~\cite{kim2024openvla} and expert-style VLAs~\cite{pi_0, intelligence2025pi_}.
At test time, the auxiliary units are discarded, allowing the original VLA architecture to run unchanged while benefiting from the integrated information about world dynamics.


Through extensive experiments on multi-task simulation benchmarks (LIBERO and RoboCasa) and real-world evaluations, we show that Pri4R consistently outperforms state-of-the-art baselines across all settings. We further analyze supervision target variants and find that 3D point track prediction most effectively encourages learning of action-world dynamics.
Finally, we ablate the core design choices of Pri4R and show that each component is necessary for learning world dynamics and improving downstream performance.


Our contributions are summarized as follows:
\begin{itemize}
    \item We propose Pri4R, which facilitates learning world dynamics by leveraging 3D point tracks as privileged supervision~\cite{Feyereisl2014ObjectLB} during training to enrich the shared representation of VLA models.
        
    \item We show that Pri4R consistently improves the performance of SOTA VLA models across simulation and real-world tasks without any inference time overhead or architectural changes.
       
    \item Through systematic analysis, we demonstrate that predicting 3D point tracks is particularly effective for learning action-world dynamics and extensively ablate each design choice of Pri4R.
\end{itemize}
 
\section{Related Work}


\boldparagraph{Vision-Language-Action models.}
Pioneered by Robotics Transformers~\cite{brohan2022rt, zitkovich2023rt}, Vision–Language–Action (VLA) models~\cite{kim2024openvla, pi_0, intelligence2025pi_, huang2023embodied, li2023vision, bu2025agibot, cen2025worldvla, li2025bridgevla, liu2025hybridvla, team2024octo} adapt large pretrained Vision–Language Models (VLMs)~\cite{llava, flamingo, chen2023pali} for robotic control by predicting actions conditioned on visual observations and language instructions, typically via action-specific heads or modules. OpenVLA~\cite{kim2024openvla} trains a 7B VLM~\cite{karamcheti2024prismatic} by autoregressively predicting discrete action tokens on the OpenX~\cite{openx} dataset. Follow-up works improve OpenVLA by reducing the model footprint (e.g., to 1B parameters)~\cite{belkhale2024minivla} or by incorporating visual trace prompting~\cite{zheng2024tracevla}. In contrast, other methods~\cite{pi_0, team2024octo, wen2025tinyvla, li2024cogact} use generative action heads (e.g., diffusion or flow matching) to produce continuous actions. More recently, OpenVLA-OFT~\cite{kim2025fine} shows that L1 regression on continuous actions with parallel decoding can both improve performance and substantially increase inference throughput. 

\boldparagraph{Improving VLAs with forecasting.} Inspired by human anticipatory reasoning, forecasting has played an important role in VLAs.
One line of work~\cite{hu2024video, bharadhwaj2024gen2act, du2023learning, wu2023unleashing, ajay2023compositional, bu2024closed} employs generative models to explicitly predict future states, images, or videos, and then conditions action policies on these predicted modalities. This can be seen as the generative model serving as a high-level planner, guiding the low-level policy to improve task achievement. 
However, these approaches are often constrained by the accuracy of the predicted signals and the additional inference-time latency they introduce. In contrast, more recent methods leverage the VLA itself to produce auxiliary signals within the same model, using them as intermediate reasoning either before or alongside action prediction at test time. Concretely, this is achieved either by first generating intermediate signals and then causally predicting actions~\cite{zhao2025cot, tian2024predictive, zhen20243d, intelligence2025pi_, zhou2025chatvla, bu2025agibot}, or by generating signals and actions jointly in parallel~\cite{zhangdreamvla, videovla}. Although effective, these auxiliary signals (e.g., language, feature embeddings, images, videos) are not expressed in the same spatiotemporal metric space as actions, and therefore provide only indirect supervision for action learning.

\boldparagraph{Point tracking for manipulation.} 2D point tracks have been actively used in robot learning~\cite{xu2024flow, yuan2024robopoint, zheng2024tracevla}, but they provide limited geometric information. With the emergence of strong 3D point tracking models~\cite{feng2025st4rtrack, xiao2025spatialtracker}, recent work has begun to adopt 3D point tracks for robotic manipulation as a structured spatiotemporal representation for downstream learning and control. In particular, point tracks have been used for policy learning~\cite{wen2023any, yin2025object, seita2023toolflownet}, reward modeling~\cite{shipoints2reward, patel2025real, xu2024flow}, and even as action representations~\cite{huang2026pointworld, niu2025pre}.  In our work, we use 3D point tracks as a privileged training signal to encourage the model to learn geometric 4D world dynamics that are directly relevant for action prediction, without requiring point tracks at inference time.
\vspace*{-5mm}


\begin{figure*}[ht]
  \centering
  \includegraphics[width=0.99\textwidth]{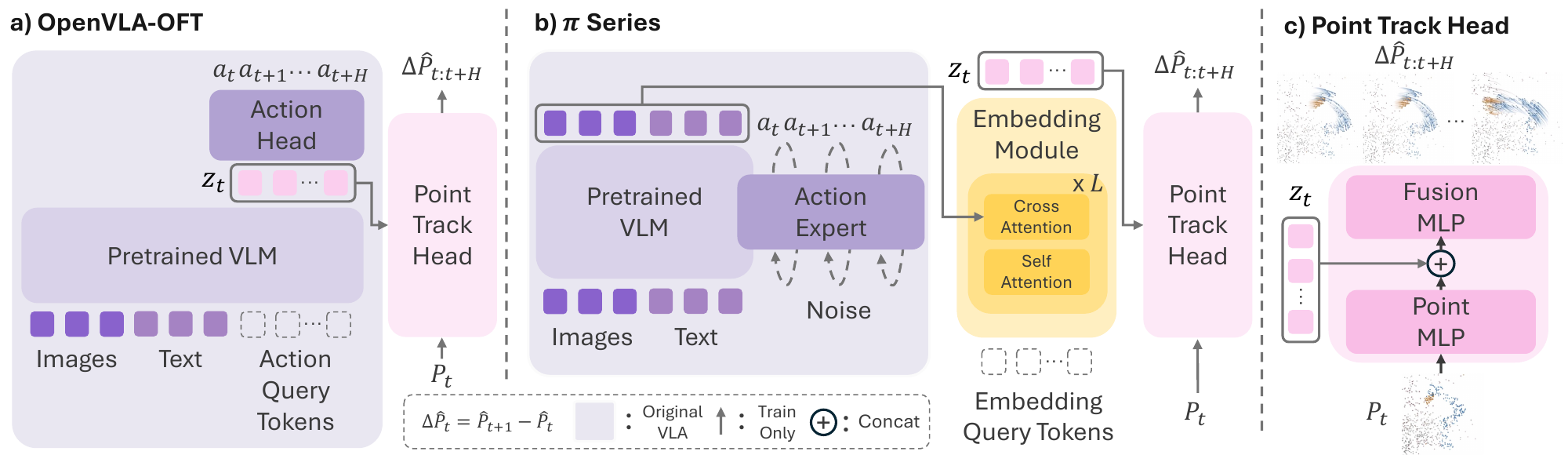}
  \caption{\textbf{Overview of Pri4R.} We augment two common VLA architectures with an auxiliary point track head that predicts per-step 3D point displacements $\widehat{\Delta P}_{t:t+H}$ from backbone embeddings $\mathbf{z}_t$ and the current point set $P_t$. (a) For backbone-centric VLAs (e.g., OpenVLA-OFT~\cite{kim2024openvla}), we set $\mathbf{z}_t$ to the final layer action-query token embeddings. (b) For expert-style VLAs (e.g., $\pi$~\cite{pi_0, intelligence2025pi_}), we condition an embedding module on the backbone’s final layer hidden states to produce $\mathbf{z}_t$. (c) The point track head encodes $P_t$ with a PointMLP, then fuses the resulting point features with $\mathbf{z}_t$ via a FusionMLP to predict future point tracks. Privileged 3D point track supervision during training forces the VLA to model how scene geometry evolves, yielding more reliable interaction and higher task success, while leaving the test-time interface and compute completely unchanged.
}
  \label{fig:Pri4R_overview}
  \vspace{-1em}
\end{figure*}

\section{Preliminaries}
\label{sec:method_prelim}
This section formalizes the Vision-Language-Action (VLA) framework and reviews representative baselines, providing the background to discuss how Pri4R incorporates world dynamics into these formulations.
\vspace{-0.5em}

\subsection{VLA Formulation}
Vision-Language-Action (VLA) models are typically developed by fine-tuning pretrained VLMs via imitation learning~\cite{learning_from_demonstration}. 
At each time step $t$, the policy receives an observation $\mathbf{o}_t \triangleq (\mathbf{I}_t, \mathbf{t}_t, \mathbf{q}_t)$ and predicts an action chunk $\mathbf{a}_{t:t+H}$ over a horizon $H$, 
where $\mathbf{I}_t=\{\mathbf{I}_t^{i}\}_{i=1}^{N}$ denotes a set of $N$ multi-view camera images, $\mathbf{t}_t$ is a tokenized language instruction, and $\mathbf{q}_t$ is the robot proprioceptive state. 
Note that, if $H=1$, the chunk reduces to a single action $\mathbf{a}_t$.

In this framework, the images $\mathbf{I}_t$ and robot state $\mathbf{q}_t$ are encoded by modality-specific encoders and projected onto the backbone's token embedding space alongside $\mathbf{t}_t$. 
Given these inputs, the VLM backbone produces high-level multi-modal embeddings. 
Since standard VLMs are not designed to output action chunks, VLAs attach an additional \emph{action-prediction component} (e.g., a lightweight action head) that consumes these embeddings to parameterize an action distribution conditioned on the backbone's internal representations (and optionally $\mathbf{q}_t$). 
The full set of model parameters $\theta$ is trained end-to-end on a demonstration dataset $\mathcal{D}$ with behavior cloning~\cite{behavior_cloning}, maximizing the log-likelihood of the demonstrated action chunks, which is given by
\begin{equation}
\max_{\theta}\; \mathbb{E}_{(\mathbf{a}_{t:t+H}, \mathbf{o}_t)\sim\mathcal{D}}
\log p_{\theta}\!\left(\mathbf{a}_{t:t+H} \mid \mathbf{o}_t\right).
\end{equation}

\subsection{VLA Baselines}
To adopt our approach, we consider two state-of-the-art VLA architectures with distinct design patterns: OpenVLA-OFT~\cite{kim2025fine} and the $\pi$ series~\cite{pi_0, intelligence2025pi_}.

OpenVLA-OFT employs an Optimized Fine-Tuning (OFT) recipe that enhances the original OpenVLA through (i) parallel decoding with bidirectional attention, (ii) action chunking, (iii) a continuous action representation, and (iv) a simple $\ell_1$ regression objective. 
Architecturally, OpenVLA-OFT replaces the discrete action-token output with an MLP regression head. This MLP head processes the final-layer hidden states at the action query token positions and maps them directly to continuous action chunks $\mathbf{a}_{t:t+H}$, enabling efficient, non-autoregressive prediction.

The $\pi$ series, in contrast, generates action chunks through a flow-matching framework. It augments the pretrained VLM backbone with a dedicated transformer action expert. 
This expert takes the robot proprioceptive state $\mathbf{q}_t$ and a noisy action chunk as inputs, predicting a corresponding vector field (velocity) for iterative denoising. The action expert integrates with the backbone by attending to its internal hidden states via shared self-attention. 
To preserve the VLM's pretrained representations, $\pi$ employs a blockwise causal attention mask; while the robotics-specific action tokens (including the noisy action chunk) can attend to the full multi-modal context, the language and image tokens are restricted from attending to these action-related tokens.

\section{Pri4R: Learning World Dynamics via Privileged 4D Representations}

Pri4R is a framework that incorporates privileged geometric information to improve the world dynamics understanding of VLA models. 
During training, we use high-fidelity 4D signals as an auxiliary supervision to refine the internal representations of the VLM backbone. 
By supervising the model to predict the physical evolution of the scene, we enable the VLA to develop a physically-aware context for robot control, without requiring any additional inputs or computational overhead during inference.


\subsection{Learning from Privileged Point Track Head}
To implement this privileged learning strategy, we augment the VLM backbone with a lightweight \emph{point track head} that predicts \textit{future 3D trajectories} of scene points. 
By feeding the backbone's multi-modal embeddings into this head, the shared representations are forced to encode the spatiotemporal constraints of the environment. 
The point track head is deliberately lightweight, consisting of two small MLPs that interface with the backbone without altering its original architecture.

At time $t$, the \emph{point MLP} embeds the current point set $P_t \in \mathbb{R}^{N_p \times 3}$ into per-point features $\mathbf{e}_t \in \mathbb{R}^{N_p \times d}$. 
To integrate scene context, we introduce a sequence of multi-modal embeddings over the action horizon, $\mathbf{z}_t = \phi(\mathbf{o}_t) \in \mathbb{R}^{H \times d}$, which is derived from the backbone $\phi$ and consumed by the action head. 
We broadcast $\mathbf{z}_t$ across points to obtain a tensor in $\mathbb{R}^{H \times N_p \times d}$ and concatenate it with the per-point features $\mathbf{e}_t$ (broadcast across the horizon $H$). 
The resulting tensor is fed into a \emph{fusion MLP} to predict per-step 3D displacements:
\begin{equation}
    \widehat{\Delta P}_{t:t+H} = \text{MLP}_{\text{fusion}}(\mathbf{z}_t \oplus \mathbf{e}_t) 
    \in \mathbb{R}^{H \times N_p \times 3},
\end{equation}
where $\oplus$ denotes the feature concatenation operator.

Through this architecture, the auxiliary loss gradients from the point track head are backpropagated into the VLM backbone, encouraging it to refine its shared representation to capture essential world dynamics. 
By jointly optimizing for both action and 3D trajectory prediction, the VLA model understands a richer, geometry-aware context that is inherently grounded in the physical evolution of the scene. 
While the core prediction mechanism remains consistent, the interface between the point track head and the VLA backbone is tailored to the specific design of each model family.
We examine the OpenVLA-OFT and $\pi$ family models; their detailed designs are illustrated in Figure~\ref{fig:Pri4R_overview} and described below.

\vspace{1mm}
\noindent\textbf{OpenVLA-OFT}: We set $\mathbf{z}_t$ to the final-layer hidden states of the action query tokens (i.e., the empty action embeddings) produced by the backbone. 
Since these are the exact embeddings mapped to actions by the action head, injecting them into the point track head allows the backbone to encode the underlying scene dynamics necessary for precise control.

\vspace{1mm}
\noindent\textbf{$\boldsymbol{\pi}$ family}: Since the action expert in $\pi$ models interacts with the backbone via masked self-attention rather than a fixed embedding, the choice of $\mathbf{z}_t$ is less straightforward. 
We introduce a lightweight transformer embedding module that takes a set of learnable query tokens and applies cross-attention over the final-layer image and language tokens from the VLM backbone. 
This yields an action-horizon embedding sequence $\mathbf{z}_t \in \mathbb{R}^{H \times d}$, which is then processed by the point track head identically to the OpenVLA-OFT case.

\begin{table*}[t]
    \centering
\caption{\textbf{Results on LIBERO.} Success rates (SRs) are reported across four task suites (500 trials per suite). Entries marked with * are taken from the OpenVLA-OFT paper~\cite{kim2025fine}. Best performances for each task are marked in \textbf{bold}. Pri4R improves average success rates of every state-of-the-art VLAs ($\pi$ series and OpenVLA-OFT) and almost every task in LIBERO.}
    \label{tab:libero} 
    \setlength\tabcolsep{15pt} 
    \scalebox{0.95}{\begin{tabular}{l|c|c|c|c|c}
        \toprule
        & Average & LIBERO - Spatial & LIBERO - Object & LIBERO - Goal & LIBERO - Long \\
        \cline{2-6}
        & SR ($\uparrow$) & SR ($\uparrow$) & SR ($\uparrow$) & SR ($\uparrow$) & SR ($\uparrow$) \\
        \midrule
        Diffusion Policy~\cite{chi2025diffusion}* & 72.4 & 78.3 & 92.5 & 68.3 & 50.5 \\
        Octo~\cite{team2024octo}* & 75.1 & 78.9 & 85.7 & 84.6 & 51.1 \\
        DiT Policy~\cite{hou2024diffusion}* & 82.4 & 84.2 & 96.3 & 85.4 & 63.8 \\
        OpenVLA~\cite{kim2024openvla}* & 76.5 & 84.7 & 88.4  & 79.2  & 53.7 \\
        \midrule
        $\pi_{0}$~\cite{pi_0}        & 87.4 $\pm$ 0.2 & 87.8 $\pm$ 0.2 & 84.9 $\pm$ 0.4 & 91.2 $\pm$ 0.9 & {85.7} $\pm$ 0.7 \\
        \rowcolor{OursRow}
    $\pi_{0}$\cite{pi_0} \textbf{+ Pri4R}            & {90.6} $\pm$ 0.2 & {92.8} $\pm$ 0.7 & {88.6} $\pm$ 0.3 & {95.3} $\pm$ 0.4 & 85.6 $\pm$ 0.2 \\
        \midrule
        $\pi_{0.5}$~\cite{intelligence2025pi_}        & 92.6 $\pm$ 0.4 & 96.1 $\pm$ 0.8 & 88.3 $\pm$ 0.8 & {95.6} $\pm$ 0.5 & 90.5 $\pm$ 1.2  \\
        \rowcolor{OursRow}
        $\pi_{0.5}$~\cite{intelligence2025pi_} \textbf{+ Pri4R}            & {94.0} $\pm$ 0.2 & \textbf{97.2} $\pm$ 0.6 & {88.9} $\pm$ 0.6 & {95.6} $\pm$ 0.5 & {94.3} $\pm$ 0.6 \\
                \midrule
        OpenVLA-OFT~\cite{kim2025fine}           & 92.7 $\pm$ 0.1 & 90.8 $\pm$ 0.3 & 98.2 $\pm$ 0.0 & 96.4 $\pm$ 0.4 & 85.5 $\pm$ 0.2 \\
        \rowcolor{OursRow}
        OpenVLA-OFT~\cite{kim2025fine}   \textbf{+ Pri4R}           & \textbf{96.3} $\pm$ 0.2 & {93.2} $\pm$ 0.4 & \textbf{98.6} $\pm$ 0.0 & \textbf{98.1} $\pm$ 0.5 & \textbf{95.3} $\pm$ 0.3 \\
        \bottomrule
    \end{tabular}}
    \vspace{-1em}
\end{table*}

\subsection{Why 3D Point Tracks as Privileged Supervision?}
Although prior works have explored a variety of predictive representations to train VLAs~\cite{zhao2025cot, tian2024predictive, zhen20243d, intelligence2025pi_, zhou2025chatvla, bu2025agibot, zhangdreamvla, videovla}, these signals are often ill-suited for learning world dynamics. 
Traditional representations are frequently (1) \textit{temporally sparse}, predicting only a goal observation at the end of the action horizon; (2) \textit{lack explicit spatial structure}, as language or latent feature embeddings do not preserve metric geometry; and (3) \textit{spatially redundant}, since dense predictions such as images or depth maps largely reproduce the input observation~\cite{zhangdreamvla}.

We therefore use \emph{3D point tracks} as the target representation. 
Compared to other representations, 3D point tracks are: (1) \textit{temporally dense}, matching the action horizon to capture fine-grained world interaction; (2) \textit{geometric}, providing metric 3D structure that promotes spatial awareness; and (3) \textit{spatially sparse}, enabling an efficient learning objective by focusing on a compact set of informative points rather than redundant dense grids like video or depth maps~\cite{zhangdreamvla}. 
Most importantly, 3D point tracks reside in the same \textit{spatiotemporal metric space} as robot actions, providing a supervisory signal that is naturally aligned with control.

\subsection{Construction of 3D Point Track Supervision}

To realize the proposed learning with a privileged information framework, we obtain 3D point tracks for every demonstration in $\mathcal{D}$ to serve as privileged supervision. 
For each training sample $(\mathbf{a}_{t:t+H}, \mathbf{o}_t)\sim\mathcal{D}$, we assume access to $N_p$ tracked points with fixed identities over the same horizon and denote the point set at time $\tau$ by
\[
P_{\tau}=\{p_j^{\tau}\}_{j=1}^{N_p},\qquad
p_j^{\tau}=(x_j^{\tau},y_j^{\tau},z_j^{\tau})\in\mathbb{R}^3,
\]
where $\tau\in\{t,\ldots,t+H+1\}$ and $j\in\{1,\ldots,N_p\}$. 
We refer to the sequence $\{P_{\tau}\}_{\tau=t}^{t+H+1}$ as the 3D point tracks for the demonstration window starting at time $t$.

Rather than directly regressing absolute positions, we supervise the point track head with per-step 3D displacements,
\[
\Delta p_j^{\tau}=p_j^{\tau+1}-p_j^{\tau},\qquad \tau\in\{t,\ldots,t+H\},
\]
and denote $\Delta P_{\tau}=\{\Delta p_j^{\tau}\}_{j=1}^{N_p}$ and $\Delta P_{t:t+H}=\{\Delta P_{\tau}\}_{\tau=t}^{t+H}$. 
Given the backbone embeddings derived from the current observation $\mathbf{o}_t$ and the current point set $P_{t}$, the point track head predicts future 3D displacements $\widehat{\Delta P}_{t:t+H}$ in parallel with the action prediction $\mathbf{a}_{t:t+H}$. 
This auxiliary branch is temporally aligned with the action horizon and utilized exclusively during training. 
By discarding the point track head after training, the final policy architecture remains identical to the original VLA, ensuring no additional computational overhead or input requirements at inference time.

\subsection{Training}
\label{sec:method_train}

\boldparagraph{Data construction.} 
We construct 3D point tracks for both simulation and real world. In simulation, we can access the ground-truth scene mesh from the simulator~\cite{mujoco}. Therefore, we initialize $N_p$ query points only at the first frame by cropping the mesh within a robot-centered 3D cube and sampling points over mesh faces. To track these points, we store the corresponding face indices and barycentric coordinates at initialization, and roll out each action sequence while retrieving the 3D locations of the same surface points at every timestep, yielding $\{P_{\tau}\}_{\tau=1}^{T}$. 
 
 For real scenes, we use an off-the-shelf 3D point tracking model \cite{xiao2025spatialtracker} to annotate pseudo-labeled tracks. To focus sampling toward foreground regions, we use a segmentation model to sample more points on robots and objects, while sampling background points uniformly. Since our real-world dataset is recorded with a fixed camera setup, we can recover stable world coordinates for the tracked points over time.

\boldparagraph{Objectives.} 
We follow the original action training objective of each VLA, $\ell_1$ regression for OpenVLA-OFT and flow matching for $\pi$ models. 
We add an auxiliary $\ell_1$ loss on 3D point track displacements $\widehat{\Delta P}_{t:t+H}$.
The total loss is
\[
\mathcal{L}
=
\mathcal{L}_{\mathrm{act}}
+
\omega_{\mathrm{pt}}\,\bigl\lVert \widehat{\Delta P}_{t:t+H}-\Delta P_{t:t+H}\bigr\rVert_{1}.
\]
where $\omega_{\mathrm{pt}}$ controls the balance between two loss terms. 

\begin{table*}[t]
    \centering
    \vspace{2mm}
    \caption{\textbf{Results on Robocasa.} Success rate (SR) for each method across 24 task suites, grouped following \cite{nasiriany2024robocasa}, and averaged over 50 trials. Pri4R improves average success rates of every state-of-the-art VLAs and almost every task in Robocasa.}
    \label{tab:robocasa} 
    \setlength\tabcolsep{3pt} 
    \scalebox{0.95}{
    \begin{tabular}{l|c|c|c|c|c|c|c|c}
        \toprule
        & Average &Pick and Place  & Open/Close doors & Open/Close drawers & Twisting knobs & Turning levers & Pressing buttons & Insertion \\
        \cline{2-9}
        & SR ($\uparrow$) & SR ($\uparrow$) & SR ($\uparrow$) & SR ($\uparrow$) & SR ($\uparrow$) & SR ($\uparrow$) & SR ($\uparrow$) & SR ($\uparrow$) \\
        \midrule
        $\pi_{0}~$\cite{pi_0}                        & 38.8 & {24.0} & 45.0 & 78.0 & 29.0 & 57.3 & 57.3 & 0.0 \\
        \rowcolor{OursRow}
        $\pi_{0}~$\cite{pi_0} \textbf{+ Pri4R}        & {42.2}~(\textcolor{blue}{+3.4}) & {24.0}~(+0.0) & {49.0}~(\textcolor{blue}{+4.0}) & {84.0}~(\textcolor{blue}{+6.0}) & {30.0}~(\textcolor{blue}{+1.0}) & {71.3}~(\textcolor{blue}{+14.0}) & {59.3}~(\textcolor{blue}{+2.0}) & {2.0}~(\textcolor{blue}{+2.0}) \\
        \midrule
        $\pi_{0.5}~$\cite{intelligence2025pi_}       & 52.9 & \textbf{54.3} & 51.0 & 75.0 & 28.0 & 79.3 & 60.0 & 4.0 \\
        \rowcolor{OursRow}
        $\pi_{0.5}$~\cite{intelligence2025pi_} \textbf{+ Pri4R}
                                                     & \textbf{57.0}~(\textcolor{blue}{+4.1})
                                                     & 52.0~(\textcolor{DecRed}{-2.3}) & \textbf{68.5}~(\textcolor{blue}{+17.5}) & \textbf{89.0}~(\textcolor{blue}{+14.0}) & \textbf{33.0}~(\textcolor{blue}{+5.0}) & \textbf{86.7}~(\textcolor{blue}{+7.4}) & 54.7~(\textcolor{DecRed}{-5.3}) & {5.0}~(\textcolor{blue}{+1.0}) \\
        \midrule
        OpenVLA-OFT~\cite{kim2025fine}               & 33.1 & 21.8 & 45.7 & 59.0 & 8.0 & 36.0 & 56.0 & 27.0 \\
        \rowcolor{OursRow}
        OpenVLA-OFT~\cite{kim2025fine}\textbf{+Pri4R} & {46.3}~(\textcolor{blue}{+13.2})
        & {23.0}~(\textcolor{blue}{+1.2}) & {61.7}~(\textcolor{blue}{+16.0}) & {80.0}~(\textcolor{blue}{+21.0}) & {25.0}~(\textcolor{blue}{+17.0}) & {66.7}~(\textcolor{blue}{+30.7}) & \textbf{79.3}~(\textcolor{blue}{+23.3}) & \textbf{34.0}~(\textcolor{blue}{+7.0}) \\
        \bottomrule
    \end{tabular}}
    \vspace{-1em}
\end{table*}

\section{Experiments}

We evaluate Pri4R's ability to inject knowledge about world dynamics during fine-tuning by using 3D point tracks as privileged supervision. Specifically, we seek to answer the following questions:

\begin{enumerate}
    \item{Does Pri4R improve state-of-the-art VLAs across diverse robot tasks?}
    \item Which characteristics of 3D point tracks are critical for world-dynamics learning and downstream control performance?
    \item How do Pri4R design choices affect the performance of the fine-tuned policy?
    \item Does Pri4R improve performance on real world tasks that require awareness about world dynamics?

\end{enumerate}

\subsection{Experimental Setup}

\boldparagraph{Baselines.}
We compare Pri4R against recent VLA and imitation-learning baselines with publicly available code: Diffusion Policy~\cite{chi2025diffusion}, Octo~\cite{team2024octo}, DiT Policy~\cite{hou2024diffusion}, OpenVLA~\cite{kim2024openvla}, OpenVLA-OFT~\cite{kim2025fine}, $\pi_0$~\cite{pi_0}, and $\pi_{0.5}$~\cite{intelligence2025pi_}.
For Pri4R, we use OpenVLA-OFT, $\pi_0$, and $\pi_{0.5}$ as backbones, as they (i) achieve SOTA results across diverse benchmarks and (ii) span two representative VLA architectures: backbone-centric policies with action heads and expert-style action models.

\boldparagraph{Implementation details.}
For OpenVLA-OFT~\cite{kim2025fine}, we apply LoRA adapters to the pretrained backbone and jointly train the point track head. We set the point supervision weight to $w_{\text{pt}}=1$ and sample $N_p=1024$ points per trajectory. All remaining model hyperparameters follow the OpenVLA-OFT default configuration. For $\pi_0$ and $\pi_{0.5}$~\cite{pi_0,intelligence2025pi_}, we use the official PyTorch implementation and fully fine-tune the base model together with the additional point track embedding module and our point track head. We set $w_{\text{pt}}=1$ and use $N_p\!=\!1024$. All other model hyperparameters follow the default settings used for the corresponding $\pi$ models. 


\subsection{Mutitask Simulation Benchmarks}

\boldparagraph{Experimental setup.}
We evaluate Pri4R on two multi-task simulation benchmarks: LIBERO~\cite{liu2023libero} and RoboCasa~\cite{nasiriany2024robocasa}.
LIBERO is a language-conditioned tabletop manipulation benchmark that tests multitask generalization across task families with varying spatial layouts, target goals, and object configurations; we report results on four suites (Spatial, Object, Goal, and Long), each with 10 tasks and 50 demonstrations per task.
RoboCasa is a large-scale kitchen manipulation benchmark with diverse scenes and articulated interactions; we evaluate 24 atomic tasks using the Human-50 dataset, which provides 50 demonstrations per task.

For both baselines, we train a single multitask policy per benchmark. Inputs include language instruction, proprioception and two camera views (thrid-person and wrist) for LIBERO and three views (left, right, wrist) for RoboCasa. For OpenVLA-OFT, we fine-tune with batch size 128 for 90k steps on LIBERO~\cite{liu2023libero}, and batch size 64 for 120k steps on RoboCasa~\cite{nasiriany2024robocasa}. For $\pi$ series, we finetune with batch size 128 for 30k steps on both benchmark. Additional implementation details are provided in the appendix.

\begin{figure}[t]
  \centering

  \includegraphics[width=0.45\textwidth]{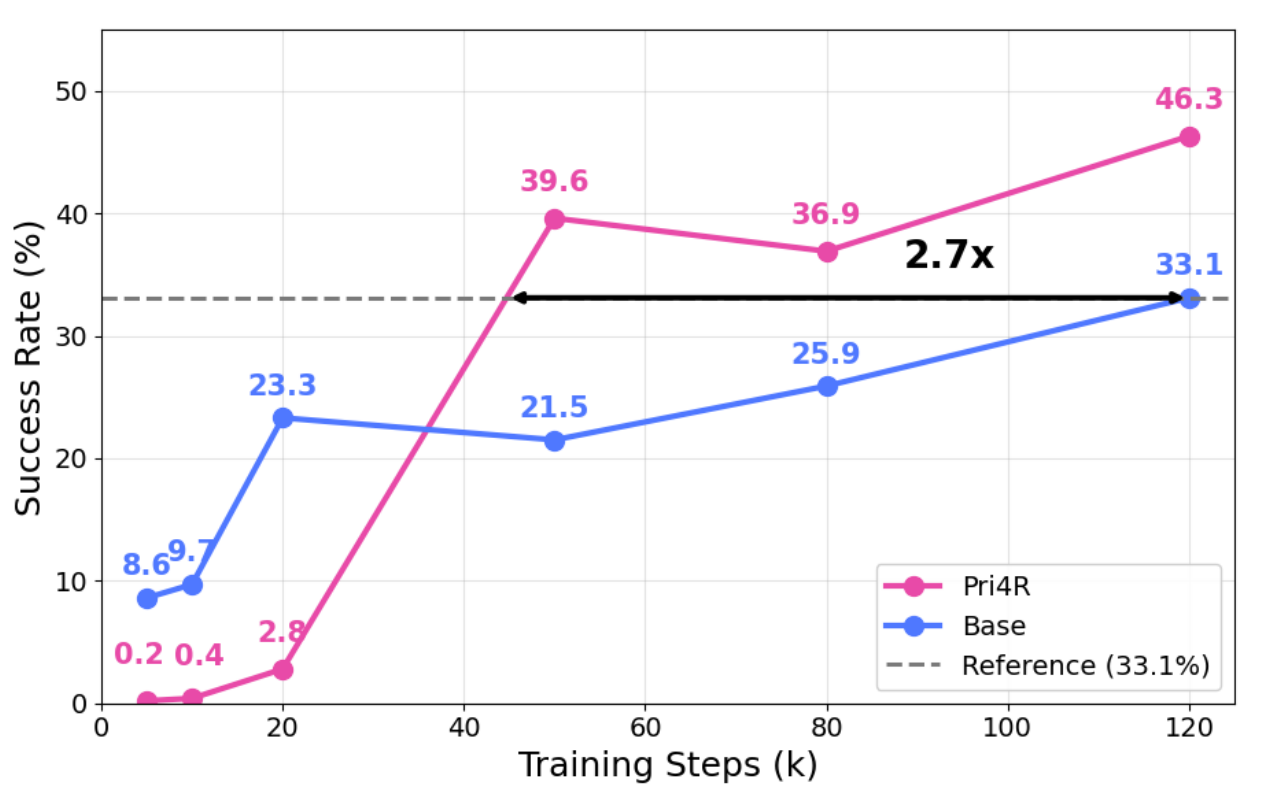}
  \caption{\textbf{Training dynamics.} Pri4R learns slowly at the early stage due to the 3D point track objective, but improves performance rapidly, reaching the baseline peak $2.7\times$ faster.}
\vspace{-1em}
  \label{fig:training_efficicency}
\end{figure}

\boldparagraph{Results.} Table~\ref{tab:libero} shows that Pri4R improves the average success rate of all strong VLA baselines on LIBERO. The gains are especially pronounced on LIBERO-Long with OpenVLA-OFT, where Pri4R yields a 9.8\% absolute improvement, indicating that privileged geometric 4D supervision encourages better modeling of action--world interactions. On the more challenging RoboCasa benchmark, which requires robust execution under randomized scene configurations, Pri4R achieves even larger success rate gains than on LIBERO (Table~\ref{tab:robocasa}), demonstrating that \emph{privileged} point track supervision transfers beyond temporally structured suites and improves generalization under stronger distribution shift. Moreover, Figure~\ref{fig:training_efficicency} illustrates Pri4R's training dynamics: performance improves more slowly during the first $\sim$20K steps due to the added point track prediction objective, but then increases rapidly, reaching the baseline's peak performance $2.7\times$ faster. This speedup corresponds to approximately 8$\times$ H200 GPU-days (about 24 hours on 8 H200s). Finally, we visualize the predicted point tracks in Figure~\ref{fig:predicted_point_tracks}. Additional results are shown in the Appendix.

\begin{table}[t]
    \centering
    \caption{\textbf{Analysis of supervision.} Average success rates on RoboCasa. 3D point track is the most effective supervision, while using both robot and scene points is crucial. }
    \label{tab:track_analysis_combined}
    \setlength{\tabcolsep}{8pt} 
    \begin{tabular}{p{0.66\columnwidth}cc}
        \toprule
        \textbf{Method} & \textbf{SR} $\uparrow$ & \textbf{$\Delta$} \\
        \midrule
        OpenVLA-OFT~\cite{kim2025fine} & 33.1 & -- \\
        \midrule
        \multicolumn{3}{l}{\textit{What supervision signal?}} \\
        \addlinespace[2pt]
        + Goal point set & 33.8 & +0.7 \\
        + 2D point track & 37.0 & +3.9 \\
        + Depth & 42.3 & +8.3 \\
        \textbf{+ Ours (3D point track)} & \textbf{46.3} & \textbf{+13.2} \\
        \midrule
        \multicolumn{3}{l}{\textit{What to track?}} \\
        \addlinespace[2pt]
        + Only scene points & 35.2 & +2.1 \\
        + Only robot points & 43.8 & +10.7 \\
        \textbf{+ Ours (both scene and robot)} & \textbf{46.3} & \textbf{+13.2} \\
        \bottomrule
    \end{tabular}
    \vspace{-1em}
\end{table}

\begin{table*}[t]
    \centering
    \caption{\textbf{Ablation on $P_t$ as input.} $\Delta$ denotes absolute improvement over the OpenVLA-OFT baseline~\cite{kim2025fine}. ``$P_t$ input'' indicates that the point set $P_t$ is appended to the backbone as additional tokens alongside image and language inputs. ``$P_{t:t+H}$ track'' refers to our proposed displacement supervision over the action horizon.}
    \label{tab:pt_input_ablation}
    \setlength{\tabcolsep}{4pt}
    \scalebox{0.95}{%
     \begin{tabular}{l|c|c|ccccccc}
        \toprule
        & Need $P_t$ at test? & \multicolumn{1}{c}{\textbf{Avg.}} & \multicolumn{1}{c}{PnP} & \multicolumn{1}{c}{Doors} & \multicolumn{1}{c}{Drawers} & \multicolumn{1}{c}{Knobs} & \multicolumn{1}{c}{Lever} & \multicolumn{1}{c}{Button} & \multicolumn{1}{c}{Insertion} \\
        \cline{2-10}
        &  & SR / $\Delta$ & SR / $\Delta$ & SR / $\Delta$ & SR / $\Delta$ & SR / $\Delta$ & SR / $\Delta$ & SR / $\Delta$ & SR / $\Delta$ \\
        \midrule
        OpenVLA-OFT~\cite{kim2025fine}
            & No
            & 33.1 / +0.0
            & 21.8 / +0.0
            & 45.7 / +0.0
            & 59.0 / +0.0
            & 8.0 / +0.0
            & 36.0 / +0.0
            & 56.0 / +0.0
            & 27.0 / +0.0 \\
        \midrule
        + $P_t$ input
            & Yes
            & 33.3 / \textcolor{blue}{+0.2}
            & 14.0 / \textcolor{DecRed}{-7.8}
            & 48.0 / \textcolor{blue}{+2.3}
            & 59.0 / +0.0
            & 10.0 / \textcolor{blue}{+2.0}
            & 49.3 / \textcolor{blue}{+13.3}
            & 59.3 / \textcolor{blue}{+3.3}
            & 26.0 / \textcolor{DecRed}{-1.0} \\
        + $P_t$ input + $P_{t:t+H}$ track
            & Yes
            & 34.5 / \textcolor{blue}{+1.4}
            & 14.3 / \textcolor{DecRed}{-7.5}
            & 49.0 / \textcolor{blue}{+3.3}
            & 59.0 / +0.0
            & 10.0 / \textcolor{blue}{+2.0}
            & 54.0 / \textcolor{blue}{+18.0}
            & 60.7 / \textcolor{blue}{+4.7}
            & 29.0 / \textcolor{blue}{+2.0} \\
        \midrule
        \rowcolor{OursRow}
        \textbf{+ $P_{t:t+H}$ track (Ours)}
            & \textbf{No}
            & \textbf{46.3 / \textcolor{blue}{+13.2}}
            & \textbf{23.0 / \textcolor{blue}{+1.2}}
            & \textbf{61.7 / \textcolor{blue}{+16.0}}
            & \textbf{80.0 / \textcolor{blue}{+21.0}}
            & \textbf{25.0 / \textcolor{blue}{+17.0}}
            & \textbf{66.7 / \textcolor{blue}{+30.7}}
            & \textbf{79.3 / \textcolor{blue}{+23.3}}
            & \textbf{34.0 / \textcolor{blue}{+7.0}} \\
        \bottomrule
    \end{tabular}}
\end{table*}
\begin{table}[t]
    \centering
    \caption{\textbf{Ablations on embedding module with $\pi_{0.5}$.} Average success rate (SR) on RoboCasa. Each design choice of our embedding module is important.}
    \label{tab:pi_ablation}
    \setlength{\tabcolsep}{12pt}
    \begin{tabular}{lcc}
        \toprule
        \textbf{Method} & \textbf{SR} $\uparrow$ & \textbf{$\Delta$} \\
        \midrule
        $\pi_{0.5}$~\cite{intelligence2025pi_} & 52.9 & -- \\
        \midrule
        \multicolumn{3}{l}{\textit{Design choice of embedding module}} \\
        \addlinespace[2pt]
        +~Point expert & 53.4 & +0.5 \\
        +~Backbone query token (attend action) & 54.4 & +1.5 \\
        +~Backbone query token & 54.8 & +1.9 \\
        \midrule
        \textbf{+~Ours} & \textbf{57.0} & \textbf{+4.1} \\
        \bottomrule
        \vspace{-1em}
    \end{tabular}
    \vspace{-3mm}
\end{table}

\subsection{Analysis of 3D Point Track Supervision}\label{subsection:Analysis of 3D Point Track Supervision}
\boldparagraph{Temporality \& Spatiality.}
To isolate the roles of temporal density and metric geometry in point track supervision, we evaluate two controlled variants of our 3D full-horizon displacement target, $\Delta P_{t:t+H}$.
For \emph{temporality}, we replace $\Delta P_{t:t+H}$ with a goal-only objective that predicts only the terminal point set $P_{t+H}$; concretely, we mean-pool backbone embeddings before the fusion MLP and keep the rest unchanged.
For \emph{spatiality}, we preserve temporal density but remove metric 3D structure by projecting 3D tracks to a single camera view and supervising with 2D tracks.
Table~\ref{tab:track_analysis_combined} shows that both goal-only prediction and 2D tracks yield only modest improvements, whereas supervising temporally dense \emph{and} metrically grounded 3D tracks over the full horizon leads to substantially larger gains, indicating that the benefits are strongest when \textbf{temporality and spatiality are jointly present} in the supervision signal.

\boldparagraph{Spatial redundancy} To assess whether 3D point tracks provide advantages over spatially dense supervision such as depth, we replace point track prediction with future depth map prediction. Because depth maps are high-resolution, we first train a depth variational autoencoder (VAE) and use its latent codes as supervision instead of raw depth. Table~\ref{tab:track_analysis_combined} shows that predicting depth improves performance over the baseline, but it consistently underperforms Pri4R with 3D point tracks. We attribute this gap to two limitations of depth supervision: (i) depth maps are spatially dense and therefore temporally redundant under a static camera, and (ii) they do not provide identity registration across time. Consequently, depth provides a weaker signal for learning how the scene evolves under interaction compared to tracking consistent 3D points over time.

\boldparagraph{Scene-robot interaction.}
Finally, we compare tracking only environment points or only robot points to our world tracking that couples both. Tracking only the robot mainly captures self-motion, while tracking only the scene weakens sensitivity to contact and articulation. Table~\ref{tab:track_analysis_combined} shows that neither matches world tracking, suggesting that modeling robot--environment interaction is key to improving performance in complex manipulation tasks.

\subsection{Ablations on Design Choices}
\boldparagraph{Effect of $P_t$ input.} A key design choice in Pri4R is to feed the current point set $P_t$ \textit{only to the point track head}, rather than to the backbone~\cite{bhat20253d} or the action head~\cite{sun2025geovla, li2025pointvla}. This has two practical advantages: (1) it requires no 3D observations at inference time, and (2) it mitigates distribution shift from pretraining, where $\mathbf{o}_t$ consists only of images and language. As shown in Table~\ref{tab:pt_input_ablation}, appending $P_t$ to the backbone can improve several suites by injecting explicit 3D cues, but it reduces performance on Pick-and-Place, where language plays a larger role~\cite{lian2026langforcebayesiandecompositionvision}. We attribute this drop to introducing a new token type into the VLM embedding space, which can perturb the pretrained representation~\cite{sun2025geovla, li2025pointvla}. Adding point track supervision on top of the backbone $P_t$ input improves the average success rate, but the PnP degradation persists. In contrast, conditioning \emph{only} the point track head on $P_t$ while supervising it with point tracks yields consistent gains across suites and even improves PnP performance. 
\begin{figure}[t]
  \centering
  \includegraphics[width=0.50\textwidth]{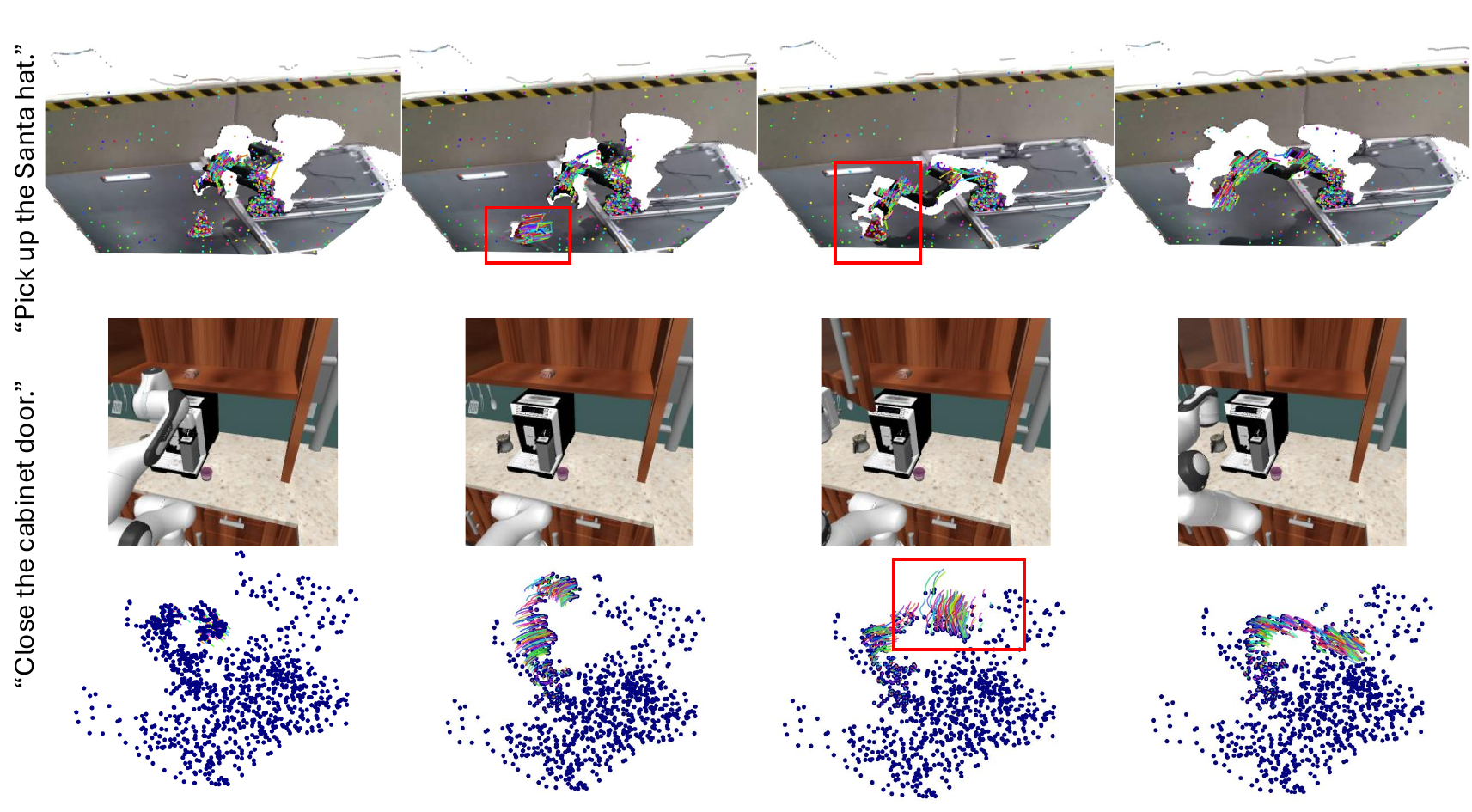}
  \caption{\textbf{Predicted point tracks in simulation and the real world.} Future trajectories are visualized in a rainbow color map. As highlighted in the red boxes, Pri4R accurately predicts point tracks for both scene elements and the robot.}
  \label{fig:predicted_point_tracks}
  \vspace{-1em}
\end{figure}

\begin{table}[t]
    \centering
\caption{\textbf{Point-track head ablations.} SR of OpenVLA-OFT on LIBERO Long. Our point-track head has two slots: \emph{Point MLP} and \emph{Fusion MLP}. We plug in different architectures for each slot and report SR; $\Delta$ is w.r.t. the no-head baseline.}
    \label{tab:pt_head_arch_ablation}
    \setlength{\tabcolsep}{14pt}
    \begin{tabular}{@{}l l r r@{}}
        \toprule
        \textbf{Point Encoder} & \textbf{Fusion Module} & \textbf{SR} $\uparrow$ & \textbf{$\Delta$} \\
        \midrule
        \multicolumn{2}{@{}l}{\textbf{OpenVLA-OFT (no point-track head)}} & 89.2 & -- \\
        \midrule
        PointNet~\cite{pointnet} & \textbf{Ours} & 80.8 & $-8.4$ \\
        PtTrans.~\cite{wu2024ptv3} & \textbf{Ours} & 92.4 & $+3.2$ \\
        \textbf{Ours} & Transformer~\cite{transformer} & 92.2 & $+3.0$ \\
        \midrule
        \textbf{Ours} & \textbf{Ours} & \textbf{94.4} & $\mathbf{+5.2}$ \\
        \bottomrule
    \end{tabular}
    \vspace{-2em}
\end{table}
\boldparagraph{Ablations on the embedding module.}
While OpenVLA-OFT only requires adding a point track head, the $\pi$ family additionally needs an embedding module to produce $\mathbf{z}_t\in\mathbb{R}^{H\times d}$.
Table~\ref{tab:pi_ablation} compares several ways of constructing $\mathbf{z}_t$.
A simple baseline introduces a \emph{point expert} mirroring the action expert, conditioned via layer-wise self-attention with the backbone.
We also evaluate learnable query tokens injected into the backbone~\cite{zhangdreamvla}, either allowing attention to action tokens or masking this interaction.
All alternatives underperform our embedding module, highlighting the importance of our design for learning world dynamics via point tracking.

\boldparagraph{Ablations on point track head. } 
We conduct an ablation study on the tracking head architecture, which consists of two components: (i) a point embedding module and (ii) a fusion module that combines action embeddings with point embeddings to predict point tracks. Table~\ref{tab:pt_head_arch_ablation} compares alternative designs, including PointNet and point-transformer point embedding, as well as a transformer-based fusion module. Overall, we find that our lightweight MLP design for both modules achieves the best performance, outperforming heavier point encoders and fusion transformers.

\subsection{Real World Evaluation}
To evaluate the Pri4R's spatiotemporal capabilities in real world settings, we conduct four real world tasks on the OMY-F3M robot, a one-arm manipulation platform.

\boldparagraph{Experimental setup}
OMY-F3M is a one-arm manipulation platform with a 6-DoF arm actuated by DYNAMIXEL-Y motors. We operate the robot at 10Hz with a 7-dimensional action space comprising six target absolute joint angles and one gripper command. Observations are collected from three camera viewpoints (top, wrist-mounted, and left third-person). We finetune OpenVLA-OFT for 60k-150k steps per task with an action chunk size of 10, and execute the full action chunk at inference time. We evaluate models on four real world tasks that require world-dynamics understanding:

\begin{enumerate}[leftmargin=*]
\item \textbf{Pick-and-place over an obstacle.} The robot must place an object over obstacles of three distinct heights. We collected 45 demonstrations and evaluated 30 trials. 
\item \textbf{Pick-and-place into a bin.} The robot picks up an object and places it into a bin. The object start location is either seen or unseen during training. We collected 45 demonstrations and evaluated 20 trials.
\item \textbf{Pick the farthest object.} The robot must pick up the object farthest from the robot among multiple candidates. Both the target object identity and object positions are randomized. We collected 48 demonstrations and evaluated 40 trials. 
\item \textbf{Pick a moving object.} The object is relocated while the robot is approaching, requiring the robot to track the moving object and grasp at the updated position. We collected 40 demonstrations and evaluated 36 trials.
\end{enumerate}






\begin{figure}[t]
  \centering
  \setlength{\tabcolsep}{3pt}
  \renewcommand{\arraystretch}{1.0}
  \begin{tabular}{cc}
    \includegraphics[width=0.46\columnwidth]{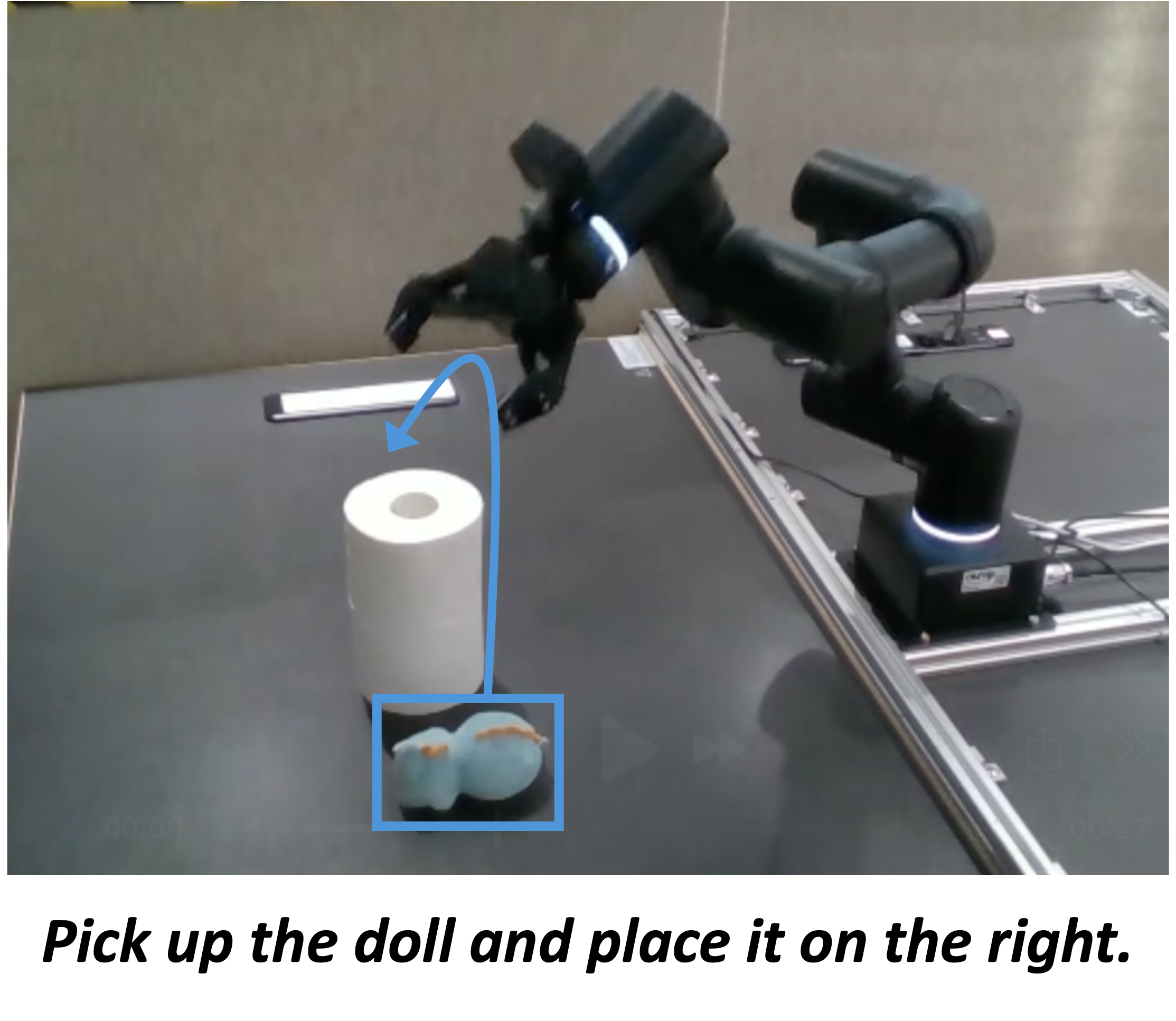} & 
    \includegraphics[width=0.46\columnwidth]{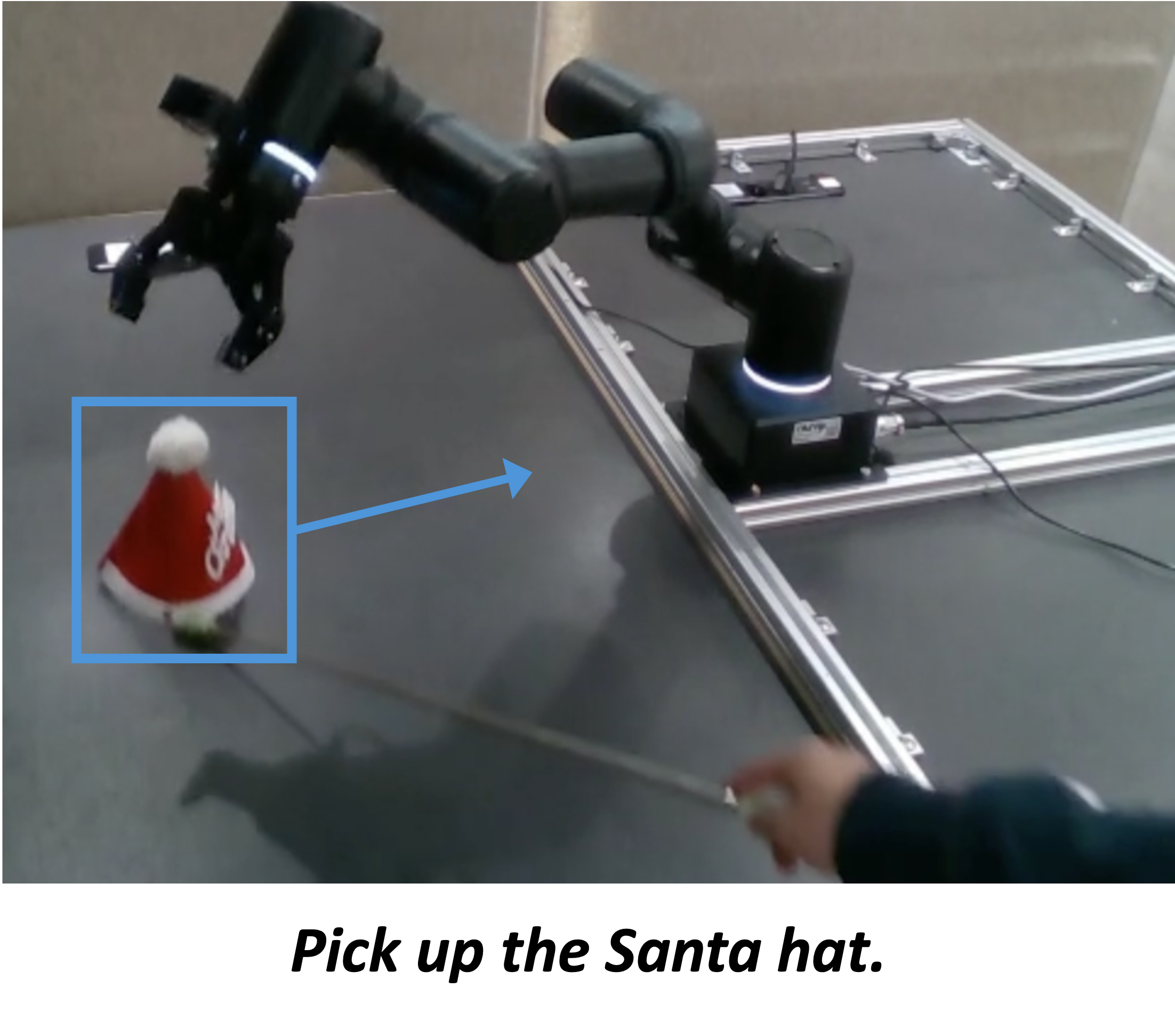} \\
    
    \includegraphics[width=0.46\columnwidth]{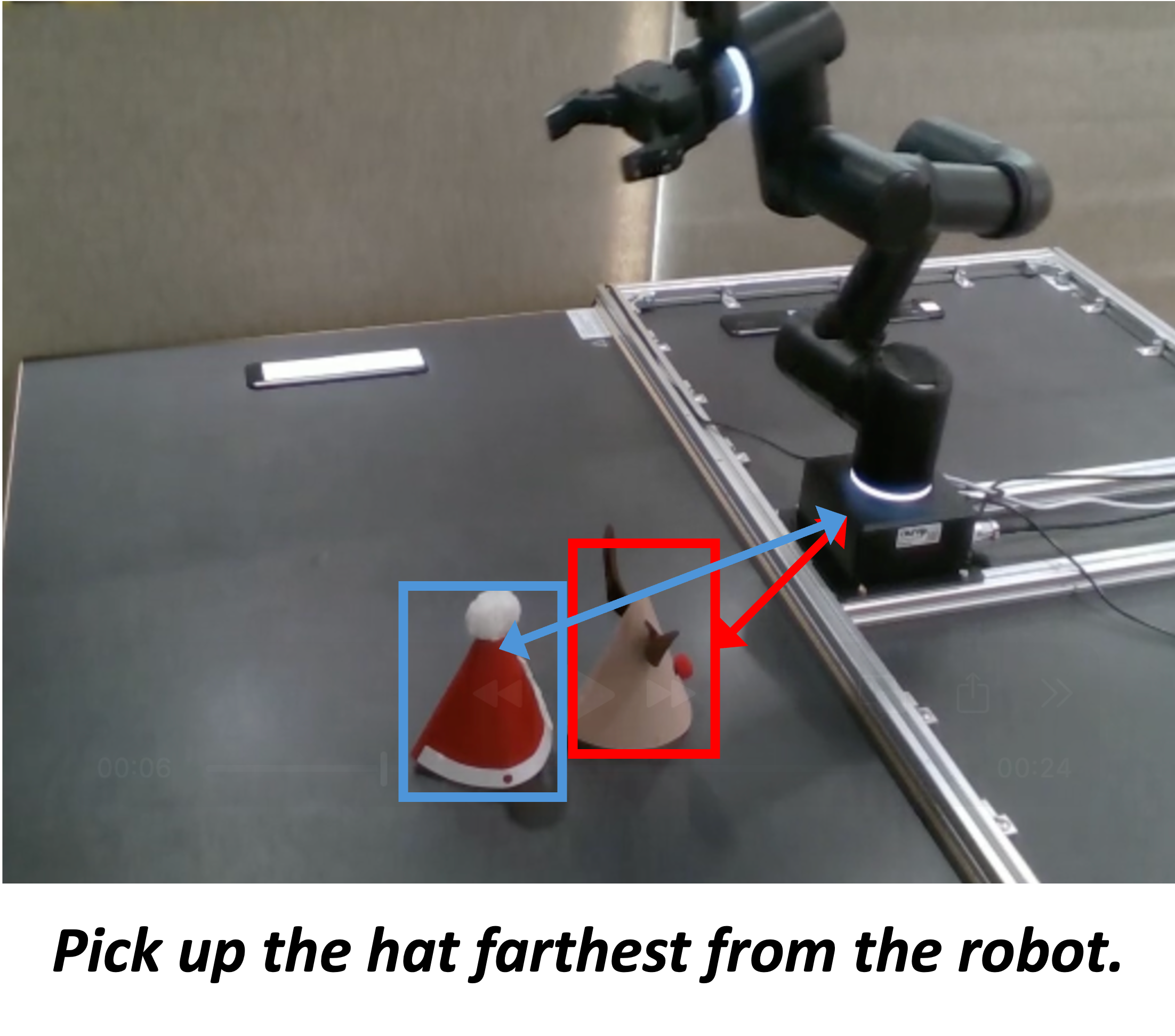} &
    \includegraphics[width=0.46\columnwidth]{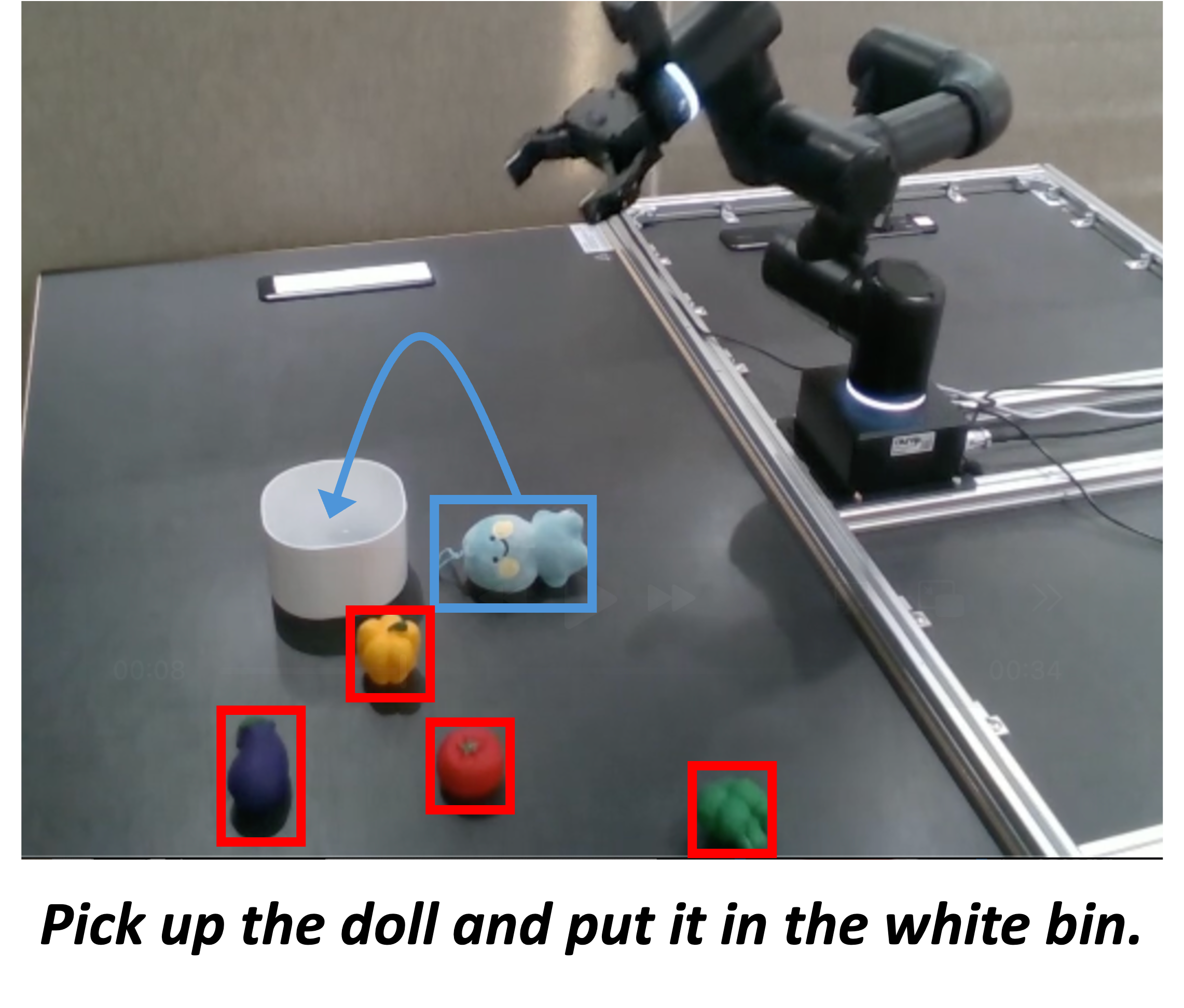} \\
  \end{tabular}  
  \vspace{-2mm}
  \caption{\textbf{Real-world setup.} Blue boxes indicate the target. In Pick the farthest object, the red box marks a closer distractor object; in Pick up the doll and place in the white bin, the red box marks a distractor.}
  \vspace*{-5mm}
  \label{fig:realworld_setup}
\end{figure}


\begin{figure}[t]
  \centering
  \setlength{\tabcolsep}{1pt}
  \renewcommand{\arraystretch}{1.0}

  \begin{tabular}{ccc}
    \rotatebox{90}{\small ~~~~Libero} &
    \includegraphics[width=0.46\columnwidth]{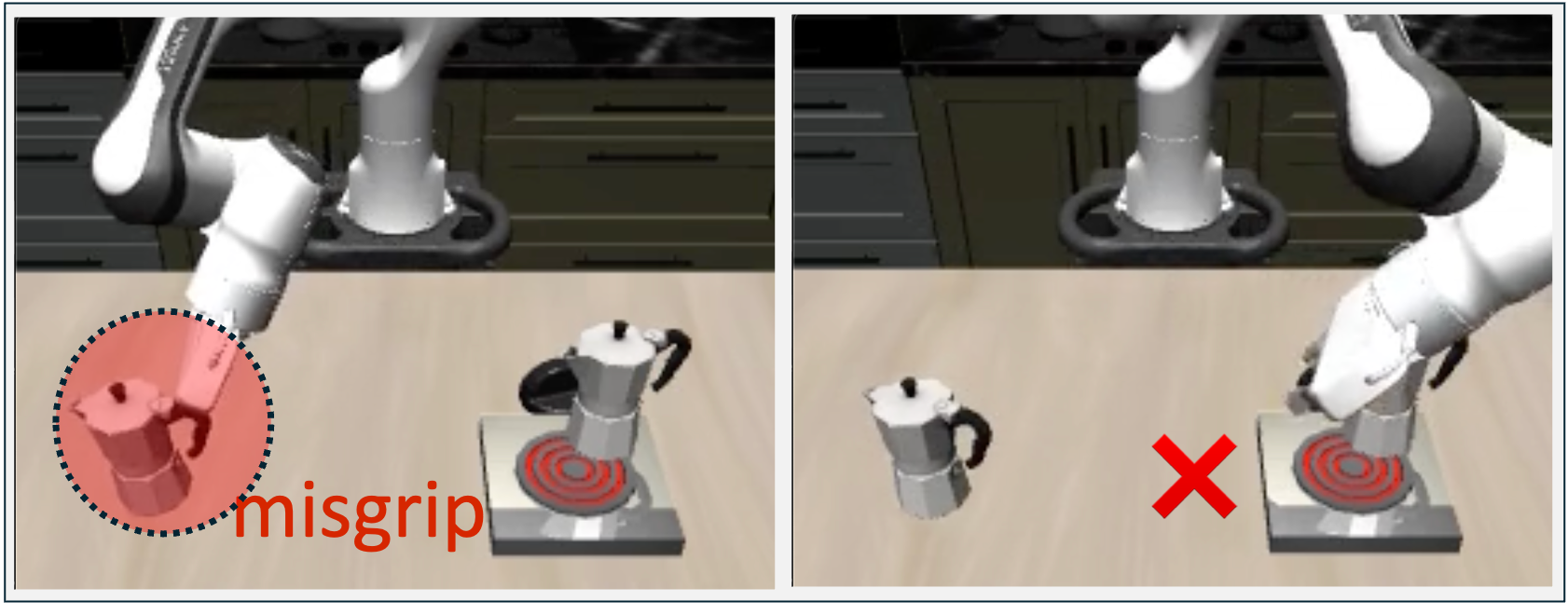} & 
    \includegraphics[width=0.46\columnwidth]{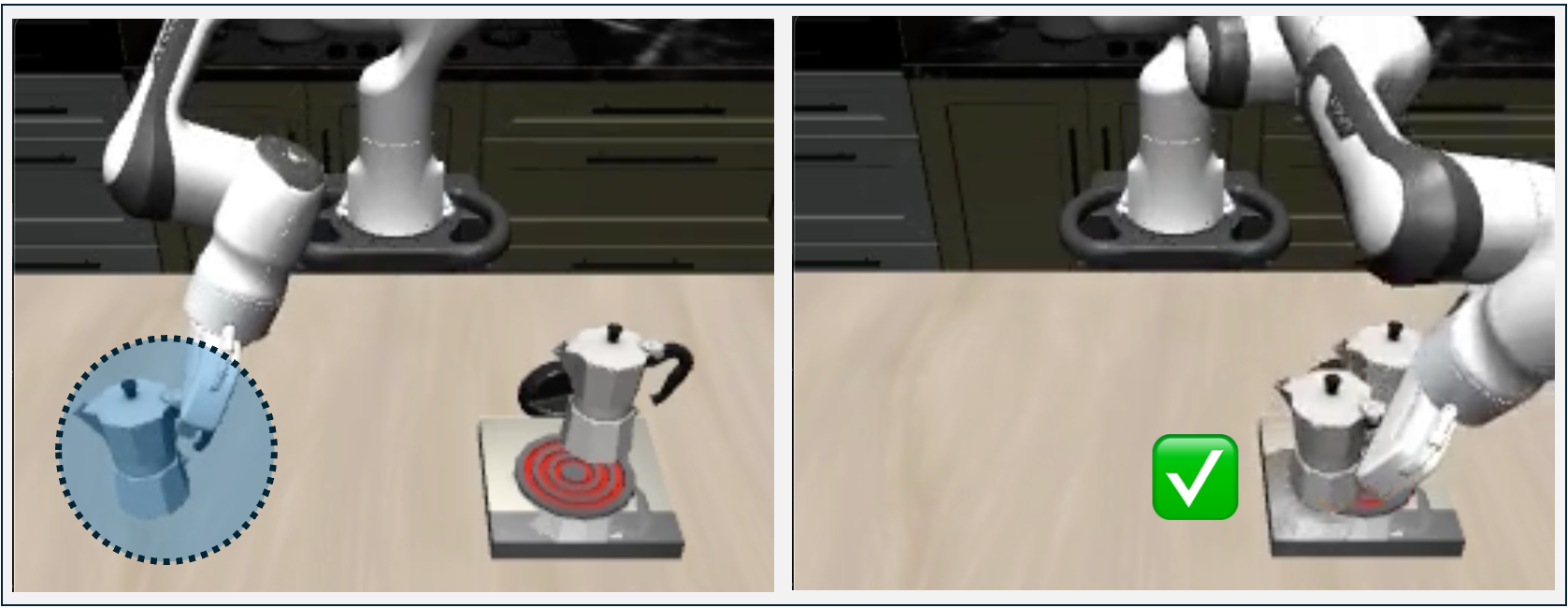} \\
    
    \rotatebox{90}{\small ~~~Robocasa} &
    \includegraphics[width=0.46\columnwidth]{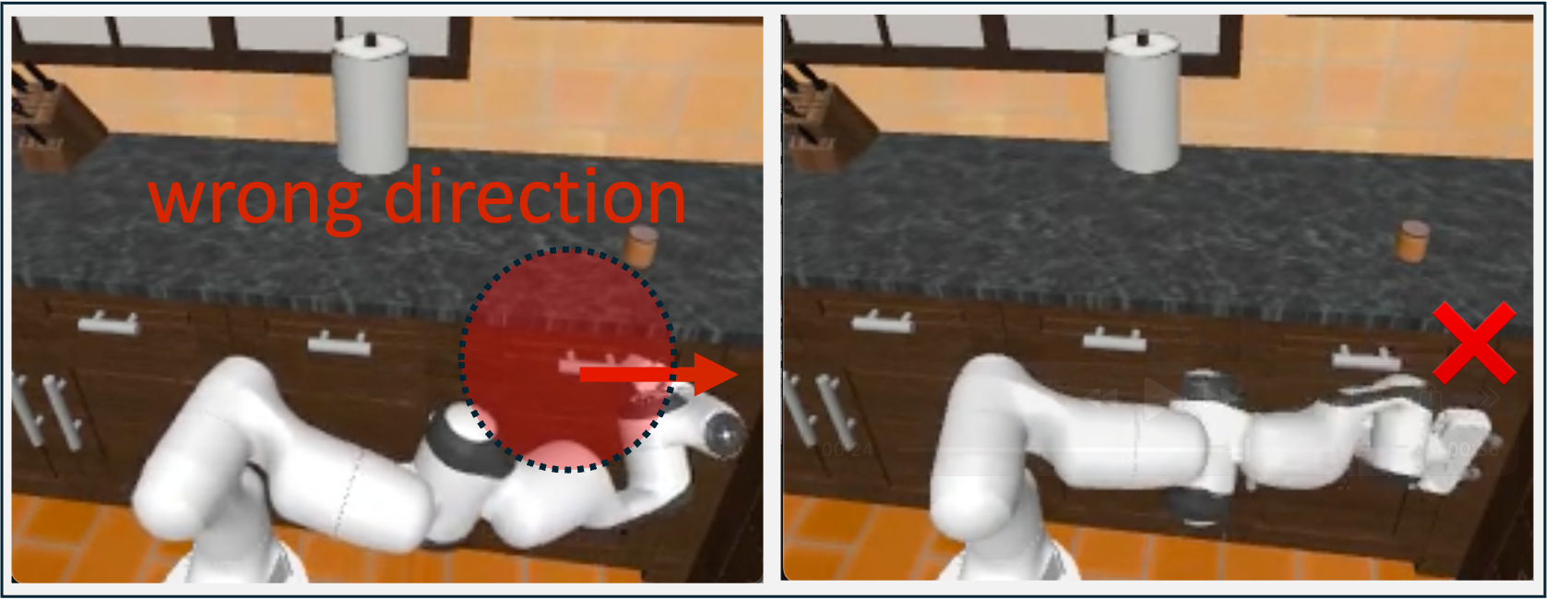} &
    \includegraphics[width=0.46\columnwidth]{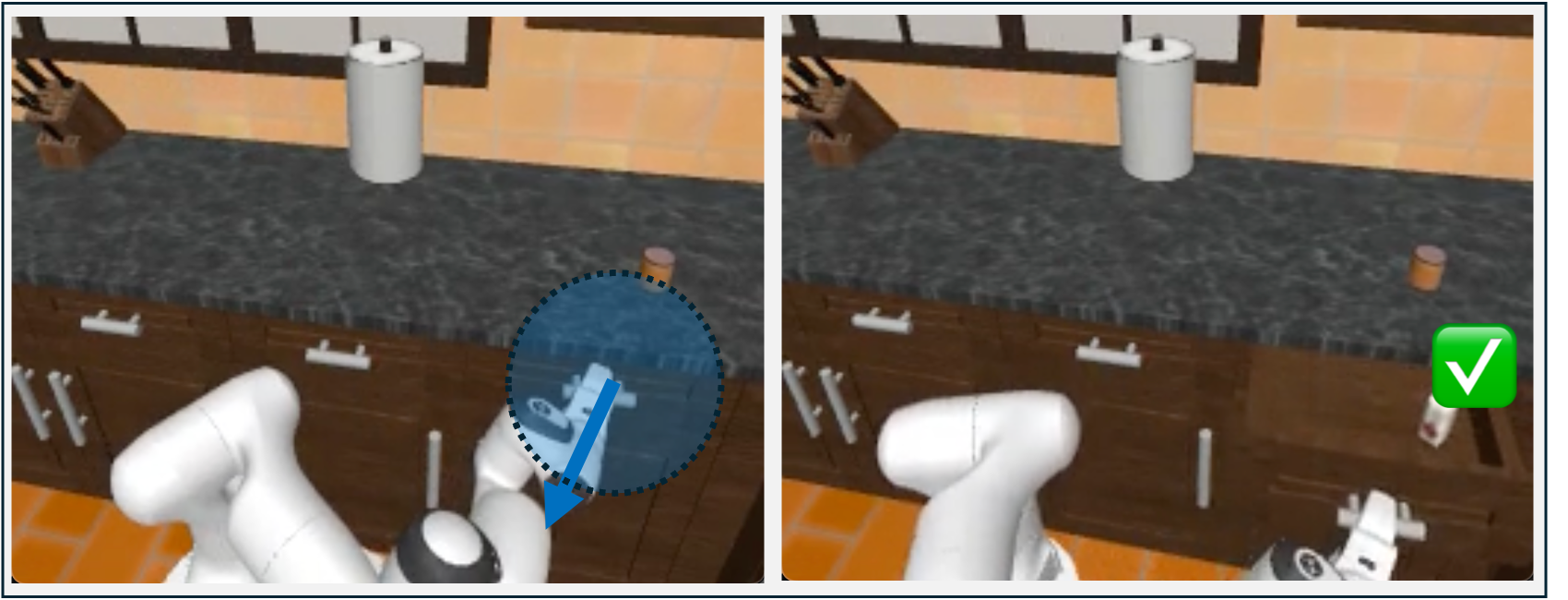} \\

    \rotatebox{90}{\small ~Real world} &
    \includegraphics[width=0.46\columnwidth]{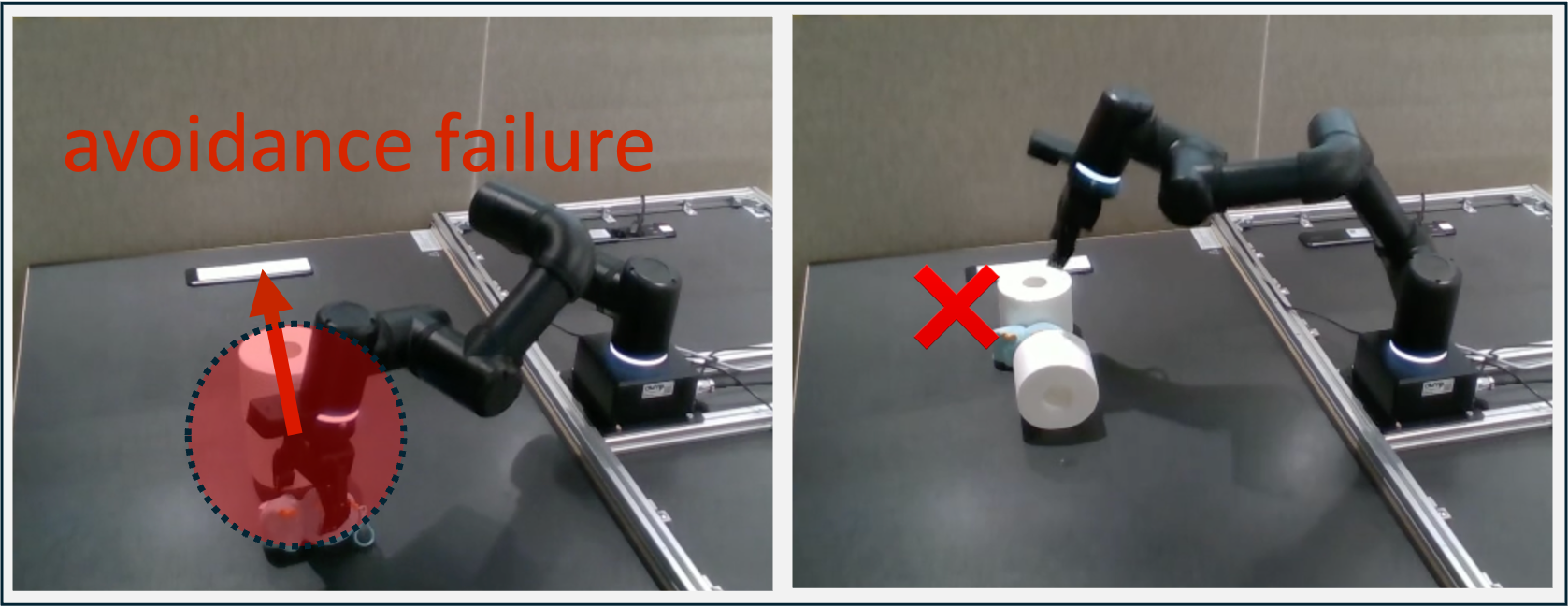} &
    \includegraphics[width=0.46\columnwidth]{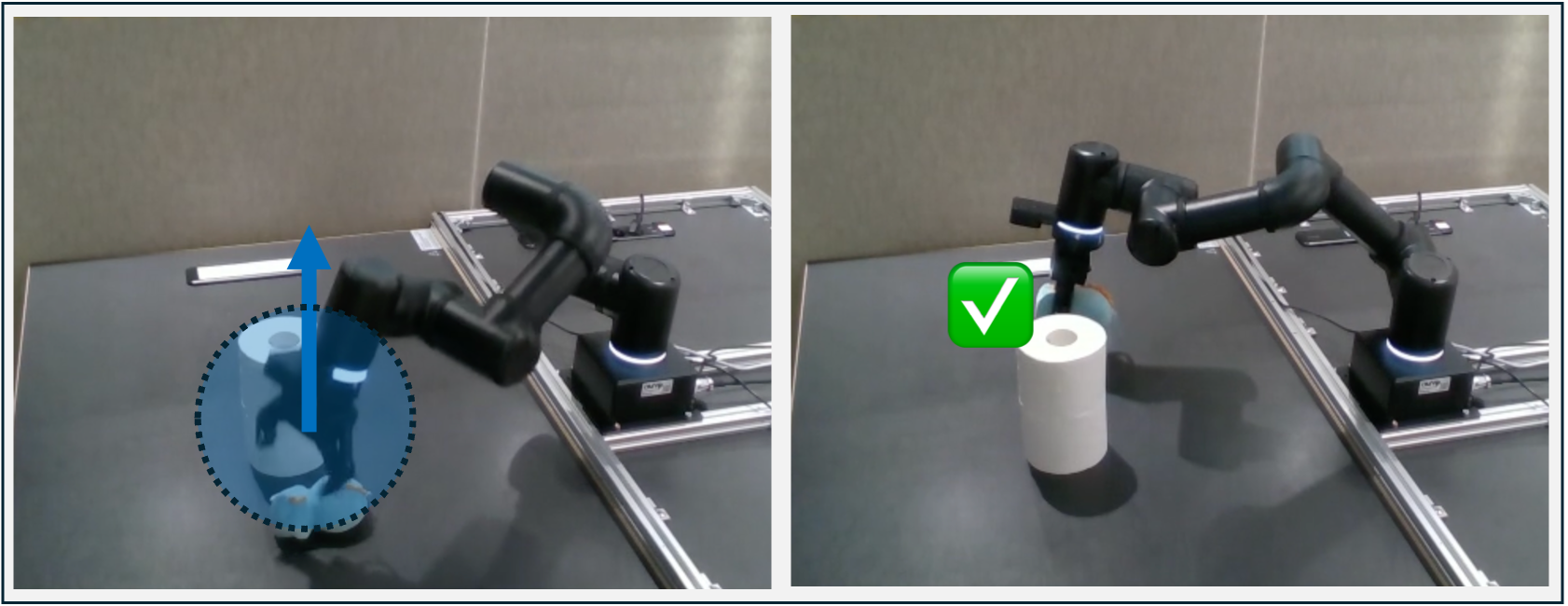} \\

    \rotatebox{90}{\small ~Real world} &
    \includegraphics[width=0.46\columnwidth]{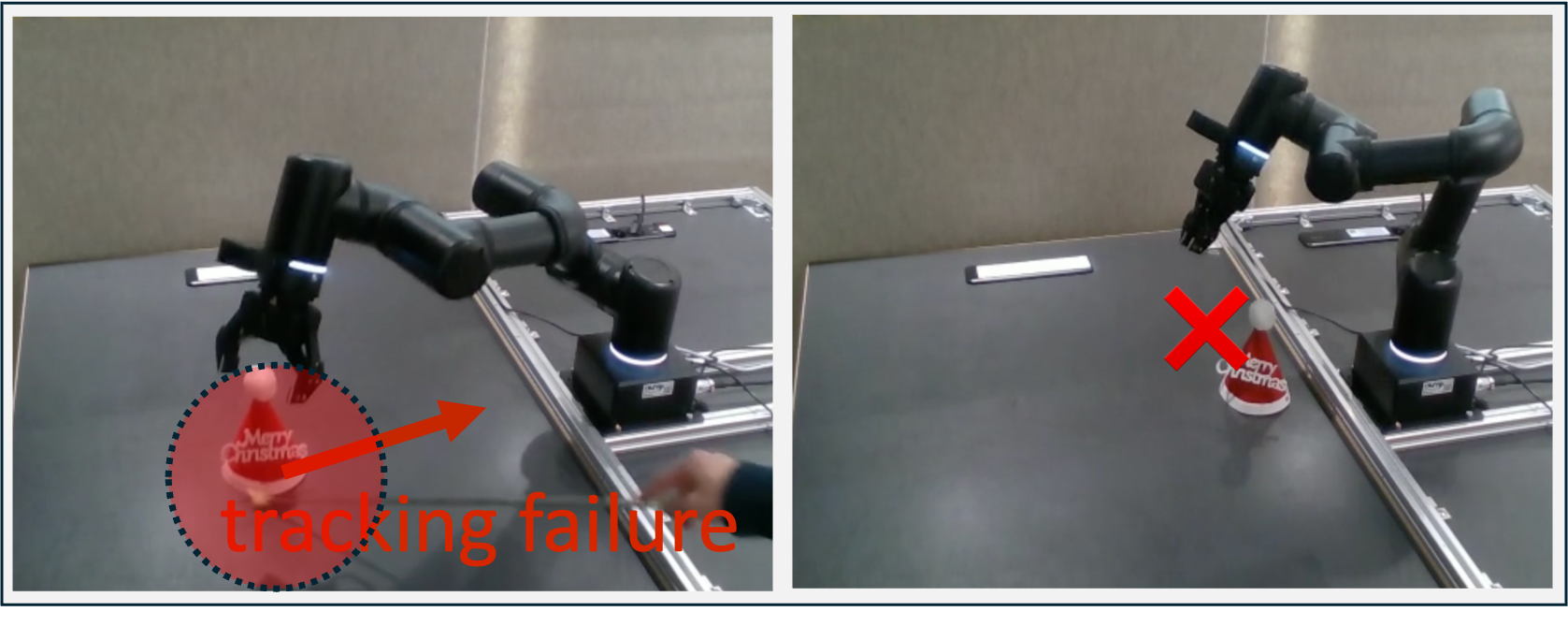} &
    \includegraphics[width=0.46\columnwidth]{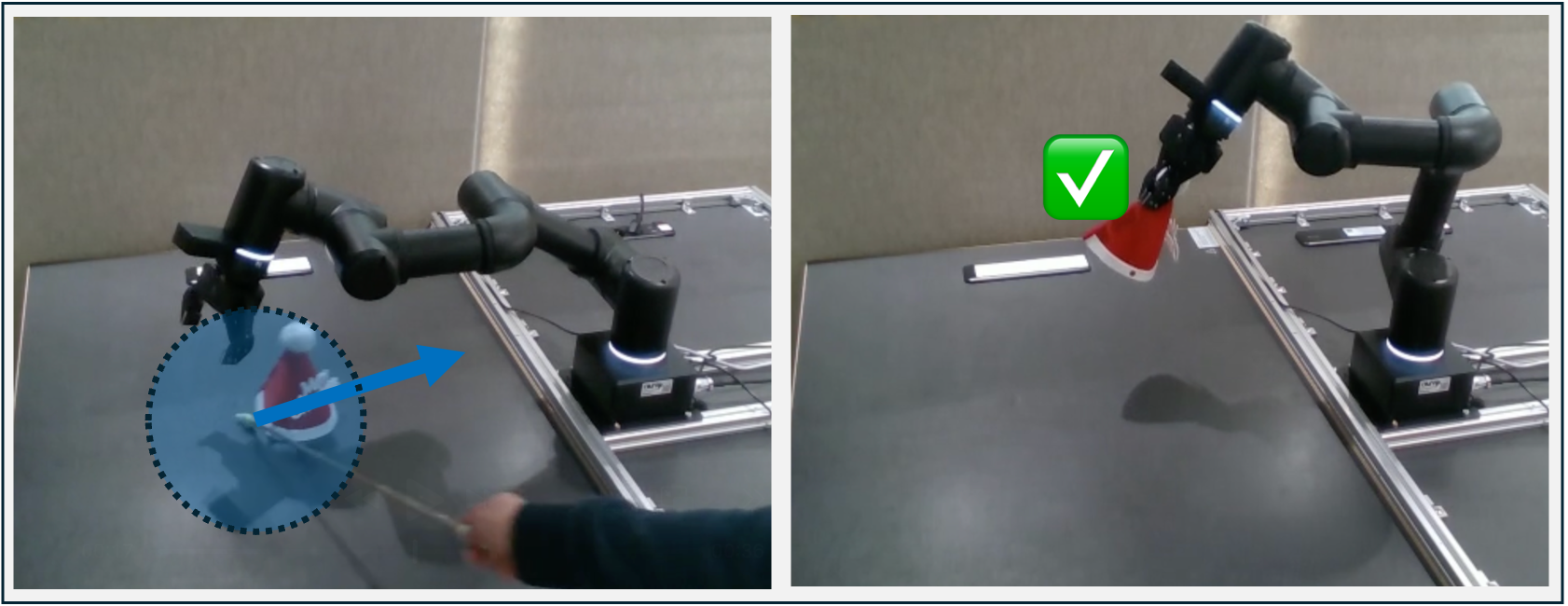} \\
    & \small (a) OpenVLA-OFT~\cite{kim2025fine} & \small (b) \textbf{OpenVLA-OFT + Pri4R}
  \end{tabular}

  \caption{\textbf{Qualitative comparison across tasks.} Baseline (left) failures vs. Our method (right) successes.
  }
  \vspace{-5pt}
  \label{fig:qual_real_world}
\end{figure}

\begin{table}[t]
    \centering
    \renewcommand{\arraystretch}{1.1}
    \setlength{\tabcolsep}{1pt}
    \caption{\textbf{Quantitative results on the real-world setup.}}
    \label{table:realworld_exp}
    \scalebox{0.95}{
        \begin{tabular}{lcccccccc}
        \toprule
        \multirow{2}{*}{Model} & Height & \multicolumn{2}{c}{Spatial} & \multicolumn{2}{c}{Depth} & \multicolumn{3}{c}{Tracking} \\
        \cmidrule(lr){2-2} \cmidrule(lr){3-4} \cmidrule(lr){5-6} \cmidrule(lr){7-9}
        & Unseen & Seen & Unseen & Seen & Unseen & Seen & Unseen & OOD \\
        \midrule
        OpenVLA-OFT & 83.3 & 100.0 & 60.0 & 45.9 & 50.0/~0.0 & ~75.0 & 41.7 & 50.0 \\
        \rowcolor{OursRow}
        \textbf{+~\textbf{Pri4R}} & 
        \textbf{96.7} & \textbf{100.0} & \textbf{80.0} & \textbf{79.2} & \textbf{50.0}/\textbf{37.5} & \textbf{100.0} & \textbf{66.7} & \textbf{66.7} \\
        \midrule
        $\pi_{0.5}$ & 60.0 & 90.0 & 30.0 & 66.7 & 50.0/~0.0 & ~50.0 & 0.0 & 33.8 \\
        \rowcolor{OursRow}
        \textbf{+~\textbf{Pri4R}} & 
        \textbf{66.7} & \textbf{100.0} & \textbf{60.0} & \textbf{95.8} & \textbf{100.0}/0.0 & \textbf{66.7} & 0.0 & \textbf{50} \\
        \bottomrule
    \vspace*{-10mm}
    \end{tabular}}
\end{table}

\boldparagraph{Results.}
Table~\ref{table:realworld_exp} shows the performance on four real world tasks designed to evaluate the understanding of world-dynamics under contact, obstacles, randomized object configurations, reaching with depth-dependent geometry, and target relocation. Pri4R consistently outperforms the OpenVLA-OFT and $\pi_{0.5}$ baselines across all tasks, demonstrating that geometric 4D supervision improves real world spatiotemporal control. As shown in Figure~\ref{fig:qual_real_world}, the baseline model tends to collide with obstacles rather than routing around them, approaches incorrect grasp locations when object positions are randomized, and in some cases closes the gripper on empty space yet still executes the remainder of the action chunk as if contact were made. In contrast, Pri4R exhibits stronger spatiotemporal and interaction awareness. It avoids collisions, re localizes targets, grasps objects from unseen locations, maintains geometrically consistent approaches, and executes grasps at the updated object position. This advantage is most evident in the moving-target setting, where Pri4R continues to update the grasp plan as the object relocates, whereas the baseline often stops early at an outdated location, reflecting limited spatiotemporal awareness. Overall, these results suggest that Pri4R transfers spatiotemporal understanding effectively to real world settings and improves robustness under distribution shifts in object placement and dynamics. More details on real world experiments are provided in the appendix.







\section{Discussion, Limitations, and Future Work}
\label{sec:discussion, limitations, and future work}

We presented Pri4R, a framework that enhances the world dynamics understanding of VLA models through privileged 4D representations. 
By supervising the model to predict 3D point tracks during training, we demonstrated that VLA backbones can develop a more physically-aware context, leading to improved control performance without any inference-time overhead. 
Our results across various benchmarks suggest that capturing the spatiotemporal evolution of a scene is a critical component for robust robot manipulation.

Despite its effectiveness, Pri4R makes no changes at inference time. While this simplicity is a strength, it may also limit performance: additional test-time computation or explicit geometric inputs could further improve robustness. Moreover, we evaluate Pri4R primarily in the fine-tuning setting on demonstration benchmarks and a small set of real-world rollouts, rather than in large-scale pretraining regimes (e.g., Embodiment X). We believe incorporating 3D point-track supervision during pretraining would have an even larger impact on learning action--world dynamics than using it only for fine-tuning. Finally, since Pri4R transfers to real-world settings, and 4D point tracks can be obtained from off-the-shelf tracking models, our approach is directly applicable to large-scale robotics datasets.

\section*{Acknowledgments}
This work was supported by LG AI Research.


\bibliographystyle{plainnat}
\bibliography{references}

@article{kim2025fine,
  title={Fine-tuning vision-language-action models: Optimizing speed and success},
  author={Kim, Moo Jin and Finn, Chelsea and Liang, Percy},
  journal={arXiv preprint arXiv:2502.19645},
  year={2025}
}

@article{nasiriany2024robocasa,
  title={Robocasa: Large-scale simulation of everyday tasks for generalist robots},
  author={Nasiriany, Soroush and Maddukuri, Abhiram and Zhang, Lance and Parikh, Adeet and Lo, Aaron and Joshi, Abhishek and Mandlekar, Ajay and Zhu, Yuke},
  journal={arXiv preprint arXiv:2406.02523},
  year={2024}
}

@article{intelligence2025pi_,
  title={$\pi_{0.5}$: a Vision-Language-Action Model with Open-World Generalization},
  author={Intelligence, Physical and Black, Kevin and Brown, Noah and Darpinian, James and Dhabalia, Karan and Driess, Danny and Esmail, Adnan and Equi, Michael and Finn, Chelsea and Fusai, Niccolo and others},
  journal={arXiv preprint arXiv:2504.16054},
  year={2025}
}

@article{liu2023libero,
  title={Libero: Benchmarking knowledge transfer for lifelong robot learning},
  author={Liu, Bo and Zhu, Yifeng and Gao, Chongkai and Feng, Yihao and Liu, Qiang and Zhu, Yuke and Stone, Peter},
  journal={Advances in Neural Information Processing Systems},
  volume={36},
  pages={44776--44791},
  year={2023}
}

@article{team2024octo,
  title={Octo: An open-source generalist robot policy},
  author={Team, Octo Model and Ghosh, Dibya and Walke, Homer and Pertsch, Karl and Black, Kevin and Mees, Oier and Dasari, Sudeep and Hejna, Joey and Kreiman, Tobias and Xu, Charles and others},
  journal={arXiv preprint arXiv:2405.12213},
  year={2024}
}

@article{kim2024openvla,
  title={Openvla: An open-source vision-language-action model},
  author={Kim, Moo Jin and Pertsch, Karl and Karamcheti, Siddharth and Xiao, Ted and Balakrishna, Ashwin and Nair, Suraj and Rafailov, Rafael and Foster, Ethan and Lam, Grace and Sanketi, Pannag and others},
  journal={arXiv preprint arXiv:2406.09246},
  year={2024}
}

@article{llava,
  title={Visual instruction tuning},
  author={Liu, Haotian and Li, Chunyuan and Wu, Qingyang and Lee, Yong Jae},
  journal={Advances in neural information processing systems},
  volume={36},
  pages={34892--34916},
  year={2023}
}

@article{flamingo,
  title={Flamingo: a visual language model for few-shot learning},
  author={Alayrac, Jean-Baptiste and Donahue, Jeff and Luc, Pauline and Miech, Antoine and Barr, Iain and Hasson, Yana and Lenc, Karel and Mensch, Arthur and Millican, Katherine and Reynolds, Malcolm and others},
  journal={Advances in neural information processing systems},
  volume={35},
  pages={23716--23736},
  year={2022}
}

@inproceedings{openx,
  title={Open x-embodiment: Robotic learning datasets and rt-x models: Open x-embodiment collaboration 0},
  author={O’Neill, Abby and Rehman, Abdul and Maddukuri, Abhiram and Gupta, Abhishek and Padalkar, Abhishek and Lee, Abraham and Pooley, Acorn and Gupta, Agrim and Mandlekar, Ajay and Jain, Ajinkya and others},
  booktitle={2024 IEEE International Conference on Robotics and Automation (ICRA)},
  pages={6892--6903},
  year={2024},
  organization={IEEE}
}

@article{droid,
  title={Droid: A large-scale in-the-wild robot manipulation dataset},
  author={Khazatsky, Alexander and Pertsch, Karl and Nair, Suraj and Balakrishna, Ashwin and Dasari, Sudeep and Karamcheti, Siddharth and Nasiriany, Soroush and Srirama, Mohan Kumar and Chen, Lawrence Yunliang and Ellis, Kirsty and others},
  journal={arXiv preprint arXiv:2403.12945},
  year={2024}
}

@inproceedings{bridge,
  title={Bridgedata v2: A dataset for robot learning at scale},
  author={Walke, Homer Rich and Black, Kevin and Zhao, Tony Z and Vuong, Quan and Zheng, Chongyi and Hansen-Estruch, Philippe and He, Andre Wang and Myers, Vivek and Kim, Moo Jin and Du, Max and others},
  booktitle={Conference on Robot Learning},
  pages={1723--1736},
  year={2023},
  organization={PMLR}
}

@article{pi_0,
  title={$\pi_0$: A Vision-Language-Action Flow Model for General Robot Control},
  author={Black, Kevin and Brown, Noah and Driess, Danny and Esmail, Adnan and Equi, Michael and Finn, Chelsea and Fusai, Niccolo and Groom, Lachy and Hausman, Karol and Ichter, Brian and others},
  journal={arXiv preprint arXiv:2410.24164},
  year={2024}
}

@article{cen2025worldvla,
  title={WorldVLA: Towards Autoregressive Action World Model},
  author={Cen, Jun and Yu, Chaohui and Yuan, Hangjie and Jiang, Yuming and Huang, Siteng and Guo, Jiayan and Li, Xin and Song, Yibing and Luo, Hao and Wang, Fan and others},
  journal={arXiv preprint arXiv:2506.21539},
  year={2025}
}

@article{li2025bridgevla,
  title={BridgeVLA: Input-Output Alignment for Efficient 3D Manipulation Learning with Vision-Language Models},
  author={Li, Peiyan and Chen, Yixiang and Wu, Hongtao and Ma, Xiao and Wu, Xiangnan and Huang, Yan and Wang, Liang and Kong, Tao and Tan, Tieniu},
  journal={arXiv preprint arXiv:2506.07961},
  year={2025}
}

@inproceedings{zhangdreamvla,
  title={DreamVLA: A Vision-Language-Action Model Dreamed with Comprehensive World Knowledge},
  author={Zhang, Wenyao and Liu, Hongsi and Qi, Zekun and Wang, Yunnan and Yu, XinQiang and Zhang, Jiazhao and Dong, Runpei and He, Jiawei and Wang, He and Zhang, Zhizheng and others},
  booktitle={The Thirty-ninth Annual Conference on Neural Information Processing Systems},
  year={2025}
}

@article{sun2025geovla,
  title={Geovla: Empowering 3d representations in vision-language-action models},
  author={Sun, Lin and Xie, Bin and Liu, Yingfei and Shi, Hao and Wang, Tiancai and Cao, Jiale},
  journal={arXiv preprint arXiv:2508.09071},
  year={2025}
}

@article{
internal_model_for_sensorimotor_integration,
author = {Daniel M. Wolpert  and Zoubin Ghahramani  and Michael I. Jordan },
title = {An Internal Model for Sensorimotor Integration},
journal = {Science},
volume = {269},
number = {5232},
pages = {1880-1882},
year = {1995}}

@article{liu2025hybridvla,
  title={Hybridvla: Collaborative diffusion and autoregression in a unified vision-language-action model},
  author={Liu, Jiaming and Chen, Hao and An, Pengju and Liu, Zhuoyang and Zhang, Renrui and Gu, Chenyang and Li, Xiaoqi and Guo, Ziyu and Chen, Sixiang and Liu, Mengzhen and others},
  journal={arXiv preprint arXiv:2503.10631},
  year={2025}
}

@inproceedings{zhao2025cot,
  title={Cot-vla: Visual chain-of-thought reasoning for vision-language-action models},
  author={Zhao, Qingqing and Lu, Yao and Kim, Moo Jin and Fu, Zipeng and Zhang, Zhuoyang and Wu, Yecheng and Li, Zhaoshuo and Ma, Qianli and Han, Song and Finn, Chelsea and others},
  booktitle={Proceedings of the Computer Vision and Pattern Recognition Conference},
  pages={1702--1713},
  year={2025}
}

@article{zheng2024tracevla,
  title={Tracevla: Visual trace prompting enhances spatial-temporal awareness for generalist robotic policies},
  author={Zheng, Ruijie and Liang, Yongyuan and Huang, Shuaiyi and Gao, Jianfeng and Daum{\'e} III, Hal and Kolobov, Andrey and Huang, Furong and Yang, Jianwei},
  journal={arXiv preprint arXiv:2412.10345},
  year={2024}
}

@article{brohan2022rt,
  title={Rt-1: Robotics transformer for real-world control at scale},
  author={Brohan, Anthony and Brown, Noah and Carbajal, Justice and Chebotar, Yevgen and Dabis, Joseph and Finn, Chelsea and Gopalakrishnan, Keerthana and Hausman, Karol and Herzog, Alex and Hsu, Jasmine and others},
  journal={arXiv preprint arXiv:2212.06817},
  year={2022}
}

@article{niu2025pre,
  title={Pre-training auto-regressive robotic models with 4d representations},
  author={Niu, Dantong and Sharma, Yuvan and Xue, Haoru and Biamby, Giscard and Zhang, Junyi and Ji, Ziteng and Darrell, Trevor and Herzig, Roei},
  journal={arXiv preprint arXiv:2502.13142},
  year={2025}
}

@article{huang2026pointworld,
  title={PointWorld: Scaling 3D World Models for In-The-Wild Robotic Manipulation},
  author={Huang, Wenlong and Chao, Yu-Wei and Mousavian, Arsalan and Liu, Ming-Yu and Fox, Dieter and Mo, Kaichun and Fei-Fei, Li},
  journal={arXiv preprint arXiv:2601.03782},
  year={2026}
}

@inproceedings{zitkovich2023rt,
  title={Rt-2: Vision-language-action models transfer web knowledge to robotic control},
  author={Zitkovich, Brianna and Yu, Tianhe and Xu, Sichun and Xu, Peng and Xiao, Ted and Xia, Fei and Wu, Jialin and Wohlhart, Paul and Welker, Stefan and Wahid, Ayzaan and others},
  booktitle={Conference on Robot Learning},
  pages={2165--2183},
  year={2023},
  organization={PMLR}
}

@article{forward_models,
title = {Forward Models for Physiological Motor Control},
journal = {Neural Networks},
volume = {9},
number = {8},
pages = {1265-1279},
year = {1996},
note = {Four Major Hypotheses in Neuroscience},
issn = {0893-6080},
author = {R.C. Miall and D.M. Wolpert}
}

@article {the_role_of_internal_models,
	author = {Flanagan, J. Randall and Wing, Alan M.},
	title = {The Role of Internal Models in Motion Planning and Control: Evidence from Grip Force Adjustments during Movements of Hand-Held Loads},
	volume = {17},
	number = {4},
	pages = {1519--1528},
	year = {1997},
	publisher = {Society for Neuroscience},
	issn = {0270-6474},
	journal = {Journal of Neuroscience}
}

@article{foward_modeling_allows_feedback,
title = {Forward modeling allows feedback control for fast reaching movements},
journal = {Trends in Cognitive Sciences},
volume = {4},
number = {11},
pages = {423-431},
year = {2000},
issn = {1364-6613},
author = {Michel Desmurget and Scott Grafton}
}

@article{sensory_input_and_control_of_grip,
author = {Johansson, Roland},
year = {1998},
month = {02},
pages = {45-59; discussion 59},
title = {Sensory Input and Control of Grip},
volume = {218},
isbn = {9780471982623},
journal = {Novartis Foundation symposium},
doi = {10.1002/9780470515563.ch4}
}

@inproceedings{xiao2025spatialtracker,
  title={SpatialTrackerV2: 3D Point Tracking Made Easy},
  author={Xiao, Yuxi and Wang, Jianyuan and Xue, Nan and Karaev, Nikita and Makarov, Iurii and Kang, Bingyi and Zhu, Xin and Bao, Hujun and Shen, Yujun and Zhou, Xiaowei},
  booktitle={ICCV},
  year={2025}
}

@article{feng2025st4rtrack,
  title={St4rtrack: Simultaneous 4d reconstruction and tracking in the world},
  author={Feng, Haiwen and Zhang, Junyi and Wang, Qianqian and Ye, Yufei and Yu, Pengcheng and Black, Michael J and Darrell, Trevor and Kanazawa, Angjoo},
  journal={arXiv preprint arXiv:2504.13152},
  year={2025}
}

@article{hu2024video,
  title={Video prediction policy: A generalist robot policy with predictive visual representations},
  author={Hu, Yucheng and Guo, Yanjiang and Wang, Pengchao and Chen, Xiaoyu and Wang, Yen-Jen and Zhang, Jianke and Sreenath, Koushil and Lu, Chaochao and Chen, Jianyu},
  journal={arXiv preprint arXiv:2412.14803},
  year={2024}
}

@article{bharadhwaj2024gen2act,
  title={Gen2act: Human video generation in novel scenarios enables generalizable robot manipulation},
  author={Bharadhwaj, Homanga and Dwibedi, Debidatta and Gupta, Abhinav and Tulsiani, Shubham and Doersch, Carl and Xiao, Ted and Shah, Dhruv and Xia, Fei and Sadigh, Dorsa and Kirmani, Sean},
  journal={arXiv preprint arXiv:2409.16283},
  year={2024}
}

@article{du2023learning,
  title={Learning universal policies via text-guided video generation},
  author={Du, Yilun and Yang, Sherry and Dai, Bo and Dai, Hanjun and Nachum, Ofir and Tenenbaum, Josh and Schuurmans, Dale and Abbeel, Pieter},
  journal={Advances in neural information processing systems},
  volume={36},
  pages={9156--9172},
  year={2023}
}

@article{wu2023unleashing,
  title={Unleashing large-scale video generative pre-training for visual robot manipulation},
  author={Wu, Hongtao and Jing, Ya and Cheang, Chilam and Chen, Guangzeng and Xu, Jiafeng and Li, Xinghang and Liu, Minghuan and Li, Hang and Kong, Tao},
  journal={arXiv preprint arXiv:2312.13139},
  year={2023}
}

@article{shi2025hi,
  title={Hi robot: Open-ended instruction following with hierarchical vision-language-action models},
  author={Shi, Lucy Xiaoyang and Ichter, Brian and Equi, Michael and Ke, Liyiming and Pertsch, Karl and Vuong, Quan and Tanner, James and Walling, Anna and Wang, Haohuan and Fusai, Niccolo and others},
  journal={arXiv preprint arXiv:2502.19417},
  year={2025}
}

@inproceedings{zhou2025chatvla,
  title={Chatvla: Unified multimodal understanding and robot control with vision-language-action model},
  author={Zhou, Zhongyi and Zhu, Yichen and Zhu, Minjie and Wen, Junjie and Liu, Ning and Xu, Zhiyuan and Meng, Weibin and Peng, Yaxin and Shen, Chaomin and Feng, Feifei and others},
  booktitle={Proceedings of the 2025 Conference on Empirical Methods in Natural Language Processing},
  pages={5377--5395},
  year={2025}
}

@article{bu2025agibot,
  title={Agibot world colosseo: A large-scale manipulation platform for scalable and intelligent embodied systems},
  author={Bu, Qingwen and Cai, Jisong and Chen, Li and Cui, Xiuqi and Ding, Yan and Feng, Siyuan and Gao, Shenyuan and He, Xindong and Hu, Xuan and Huang, Xu and others},
  journal={arXiv preprint arXiv:2503.06669},
  year={2025}
}

@article{tian2024predictive,
  title={Predictive inverse dynamics models are scalable learners for robotic manipulation},
  author={Tian, Yang and Yang, Sizhe and Zeng, Jia and Wang, Ping and Lin, Dahua and Dong, Hao and Pang, Jiangmiao},
  journal={arXiv preprint arXiv:2412.15109},
  year={2024}
}

@article{chen2023pali,
  title={Pali-x: On scaling up a multilingual vision and language model},
  author={Chen, Xi and Djolonga, Josip and Padlewski, Piotr and Mustafa, Basil and Changpinyo, Soravit and Wu, Jialin and Ruiz, Carlos Riquelme and Goodman, Sebastian and Wang, Xiao and Tay, Yi and others},
  journal={arXiv preprint arXiv:2305.18565},
  year={2023}
}

@article{li2023vision,
  title={Vision-language foundation models as effective robot imitators},
  author={Li, Xinghang and Liu, Minghuan and Zhang, Hanbo and Yu, Cunjun and Xu, Jie and Wu, Hongtao and Cheang, Chilam and Jing, Ya and Zhang, Weinan and Liu, Huaping and others},
  journal={arXiv preprint arXiv:2311.01378},
  year={2023}
}

@article{huang2023embodied,
  title={An embodied generalist agent in 3d world},
  author={Huang, Jiangyong and Yong, Silong and Ma, Xiaojian and Linghu, Xiongkun and Li, Puhao and Wang, Yan and Li, Qing and Zhu, Song-Chun and Jia, Baoxiong and Huang, Siyuan},
  journal={arXiv preprint arXiv:2311.12871},
  year={2023}
}

@article{li2024cogact,
  title={Cogact: A foundational vision-language-action model for synergizing cognition and action in robotic manipulation},
  author={Li, Qixiu and Liang, Yaobo and Wang, Zeyu and Luo, Lin and Chen, Xi and Liao, Mozheng and Wei, Fangyun and Deng, Yu and Xu, Sicheng and Zhang, Yizhong and others},
  journal={arXiv preprint arXiv:2411.19650},
  year={2024}
}

@article{wen2025tinyvla,
  title={Tinyvla: Towards fast, data-efficient vision-language-action models for robotic manipulation},
  author={Wen, Junjie and Zhu, Yichen and Li, Jinming and Zhu, Minjie and Tang, Zhibin and Wu, Kun and Xu, Zhiyuan and Liu, Ning and Cheng, Ran and Shen, Chaomin and others},
  journal={IEEE Robotics and Automation Letters},
  year={2025},
  publisher={IEEE}
}

@article{li2025pointvla,
  title={Pointvla: Injecting the 3d world into vision-language-action models},
  author={Li, Chengmeng and Wen, Junjie and Peng, Yan and Peng, Yaxin and Feng, Feifei and Zhu, Yichen},
  journal={arXiv preprint arXiv:2503.07511},
  year={2025}
}

@article{bhat20253d,
  title={3D CAVLA: Leveraging Depth and 3D Context to Generalize Vision Language Action Models for Unseen Tasks},
  author={Bhat, Vineet and Lan, Yu-Hsiang and Krishnamurthy, Prashanth and Karri, Ramesh and Khorrami, Farshad},
  journal={arXiv preprint arXiv:2505.05800},
  year={2025}
}

@misc{wang2025vggtvisualgeometrygrounded,
      title={VGGT: Visual Geometry Grounded Transformer}, 
      author={Jianyuan Wang and Minghao Chen and Nikita Karaev and Andrea Vedaldi and Christian Rupprecht and David Novotny},
      year={2025},
      eprint={2503.11651},
      archivePrefix={arXiv},
      primaryClass={cs.CV}
}

@article{zhen20243d,
  title={3d-vla: A 3d vision-language-action generative world model},
  author={Zhen, Haoyu and Qiu, Xiaowen and Chen, Peihao and Yang, Jincheng and Yan, Xin and Du, Yilun and Hong, Yining and Gan, Chuang},
  journal={arXiv preprint arXiv:2403.09631},
  year={2024}
}

@article{
    videovla,
    title={VideoVLA: Video Generators Can Be Generalizable Robot Manipulators},
    author={Yichao Shen and Fangyun Wei and Zhiying Du and Yaobo Liang and Yan Lu and Jiaolong Yang and Nanning Zheng and Baining Guo},
    booktitle={The Thirty-ninth Annual Conference on Neural Information Processing Systems(NeurIPS2025)},
    year={2025}
}

@article{wen2023any,
  title={Any-point trajectory modeling for policy learning},
  author={Wen, Chuan and Lin, Xingyu and So, John and Chen, Kai and Dou, Qi and Gao, Yang and Abbeel, Pieter},
  journal={arXiv preprint arXiv:2401.00025},
  year={2023}
}

@article{behavior_cloning,
  title={Alvinn: An autonomous land vehicle in a neural network},
  author={Pomerleau, Dean A},
  journal={Advances in neural information processing systems},
  volume={1},
  year={1988}
}

@article{learning_from_demonstration,
author = {Brenna Argall and Sonia Chernova and Manuela Veloso and Brett Browning},
title = {A survey of robot learning from demonstration},
journal = {Robotics and Autonomous Systems},
volume = {57},
number = {5},
pages = {469-483},
year = {2009},
issn = {0921-8890}
}

@article{ajay2023compositional,
  title={Compositional foundation models for hierarchical planning},
  author={Ajay, Anurag and Han, Seungwook and Du, Yilun and Li, Shuang and Gupta, Abhi and Jaakkola, Tommi and Tenenbaum, Josh and Kaelbling, Leslie and Srivastava, Akash and Agrawal, Pulkit},
  journal={Advances in Neural Information Processing Systems},
  volume={36},
  pages={22304--22325},
  year={2023}
}

@article{bu2024closed,
  title={Closed-loop visuomotor control with generative expectation for robotic manipulation},
  author={Bu, Qingwen and Zeng, Jia and Chen, Li and Yang, Yanchao and Zhou, Guyue and Yan, Junchi and Luo, Ping and Cui, Heming and Ma, Yi and Li, Hongyang},
  journal={Advances in Neural Information Processing Systems},
  volume={37},
  pages={139002--139029},
  year={2024}
}

@article{shipoints2reward,
  title={Points2Reward: Robotic Manipulation Rewards from Just One Video},
  author={Shi, Junyao and Smith, Joshua and Qian, Jianing and Jayaraman, Dinesh}
}

@article{patel2025real,
  title={A real-to-sim-to-real approach to robotic manipulation with VLM-generated iterative keypoint rewards},
  author={Patel, Shivansh and Yin, Xinchen and Huang, Wenlong and Garg, Shubham and Nayyeri, Hooshang and Fei-Fei, Li and Lazebnik, Svetlana and Li, Yunzhu},
  journal={arXiv preprint arXiv:2502.08643},
  year={2025}
}

@article{yin2025object,
  title={Object-centric 3D Motion Field for Robot Learning from Human Videos},
  author={Yin, Zhao-Heng and Yang, Sherry and Abbeel, Pieter},
  journal={arXiv preprint arXiv:2506.04227},
  year={2025}
}

@inproceedings{seita2023toolflownet,
  title={Toolflownet: Robotic manipulation with tools via predicting tool flow from point clouds},
  author={Seita, Daniel and Wang, Yufei and Shetty, Sarthak J and Li, Edward Yao and Erickson, Zackory and Held, David},
  booktitle={Conference on Robot Learning},
  pages={1038--1049},
  year={2023},
  organization={PMLR}
}

@article{yuan2024robopoint,
  title={Robopoint: A vision-language model for spatial affordance prediction for robotics},
  author={Yuan, Wentao and Duan, Jiafei and Blukis, Valts and Pumacay, Wilbert and Krishna, Ranjay and Murali, Adithyavairavan and Mousavian, Arsalan and Fox, Dieter},
  journal={arXiv preprint arXiv:2406.10721},
  year={2024}
}

@article{xu2024flow,
  title={Flow as the cross-domain manipulation interface},
  author={Xu, Mengda and Xu, Zhenjia and Xu, Yinghao and Chi, Cheng and Wetzstein, Gordon and Veloso, Manuela and Song, Shuran},
  journal={arXiv preprint arXiv:2407.15208},
  year={2024}
}

@INPROCEEDINGS{mujoco,
  author={Todorov, Emanuel and Erez, Tom and Tassa, Yuval},
  booktitle={2012 IEEE/RSJ International Conference on Intelligent Robots and Systems}, 
  title={MuJoCo: A physics engine for model-based control}, 
  year={2012},
  volume={},
  number={},
  pages={5026-5033}
}

@inproceedings{karamcheti2024prismatic,
  title={Prismatic vlms: Investigating the design space of visually-conditioned language models},
  author={Karamcheti, Siddharth and Nair, Suraj and Balakrishna, Ashwin and Liang, Percy and Kollar, Thomas and Sadigh, Dorsa},
  booktitle={Forty-first International Conference on Machine Learning},
  year={2024}
}

@misc{belkhale2024minivla,
      title={MiniVLA: A Better VLA with a Smaller Footprint}, 
      author={Suneel Belkhale and Dorsa Sadigh},
      url={https://github.com/Stanford-ILIAD/openvla-mini},
      year={2024},
}

@misc{lian2026langforcebayesiandecompositionvision,
      title={LangForce: Bayesian Decomposition of Vision Language Action Models via Latent Action Queries}, 
      author={Shijie Lian and Bin Yu and Xiaopeng Lin and Laurence T. Yang and Zhaolong Shen and Changti Wu and Yuzhuo Miao and Cong Huang and Kai Chen},
      year={2026},
      eprint={2601.15197},
      archivePrefix={arXiv},
      primaryClass={cs.AI}
}

@article{chi2025diffusion,
  title={Diffusion policy: Visuomotor policy learning via action diffusion},
  author={Chi, Cheng and Xu, Zhenjia and Feng, Siyuan and Cousineau, Eric and Du, Yilun and Burchfiel, Benjamin and Tedrake, Russ and Song, Shuran},
  journal={The International Journal of Robotics Research},
  volume={44},
  number={10-11},
  pages={1684--1704},
  year={2025},
  publisher={Sage Publications Sage UK: London, England}
}

@article{hou2024diffusion,
  title={Diffusion transformer policy},
  author={Hou, Zhi and Zhang, Tianyi and Xiong, Yuwen and Pu, Hengjun and Zhao, Chengyang and Tong, Ronglei and Qiao, Yu and Dai, Jifeng and Chen, Yuntao},
  journal={arXiv preprint arXiv:2410.15959},
  year={2024}
}

@INPROCEEDINGS{pointnet,
  author={Charles, R. Qi and Su, Hao and Kaichun, Mo and Guibas, Leonidas J.},
  booktitle={2017 IEEE Conference on Computer Vision and Pattern Recognition (CVPR)}, 
  title={PointNet: Deep Learning on Point Sets for 3D Classification and Segmentation}, 
  year={2017},
  volume={},
  number={},
  pages={77-85}
}

@inproceedings{wu2024ptv3,
    title={Point Transformer V3: Simpler, Faster, Stronger},
    author={Wu, Xiaoyang and Jiang, Li and Wang, Peng-Shuai and Liu, Zhijian and Liu, Xihui and Qiao, Yu and Ouyang, Wanli and He, Tong and Zhao, Hengshuang},
    booktitle={CVPR},
    year={2024}
}

@inproceedings{transformer,
author = {Vaswani, Ashish and Shazeer, Noam and Parmar, Niki and Uszkoreit, Jakob and Jones, Llion and Gomez, Aidan N. and Kaiser, \L{}ukasz and Polosukhin, Illia},
title = {Attention is all you need},
year = {2017},
isbn = {9781510860964},
publisher = {Curran Associates Inc.},
address = {Red Hook, NY, USA},
booktitle = {Proceedings of the 31st International Conference on Neural Information Processing Systems},
pages = {6000–6010},
numpages = {11},
location = {Long Beach, California, USA},
series = {NIPS'17}
}

@inproceedings{Feyereisl2014ObjectLB,
  title={Object Localization based on Structural SVM using Privileged Information},
  author={Jan Feyereisl and Suha Kwak and Jeany Son and Bohyung Han},
  booktitle={Advances in Neural Information Processing Systems},
  pages={208--216},
  year={2014}
}

@misc{spatialforcing2025,
      title={Spatial Forcing: Implicit Spatial Representation Alignment for Vision-language-action Model}, 
      author={Fuhao Li and Wenxuan Song and Han Zhao and Jingbo Wang and Pengxiang Ding and Donglin Wang and Long Zeng and Haoang Li},
      year={2025},
      eprint={2510.12276},
      archivePrefix={arXiv},
      primaryClass={cs.RO},
      url={https://arxiv.org/abs/2510.12276}, 
}

\clearpage
\appendices
\renewcommand{\thesection}{S.\Roman{section}}
\renewcommand{\thefigure}{S\arabic{figure}}
\renewcommand{\thetable}{S\arabic{table}}
\setcounter{figure}{0}
\setcounter{table}{0}

\twocolumn[{
\centering
{\LARGE \textbf{Pri4R: Learning World Dynamics for Vision-Language-Action Models with Privileged 4D Representation}\\[2mm]
\Large Appendix\par}
\vspace{6mm}
}]


In this Appendix, we additionally provide,
\begin{itemize}
    \item \textbf{S.I. Real World Experiment Details}
    \begin{itemize}
        \item A. Real World Setup
        \item B. Real World Experiments Details
        \item C. Hyperparameters and Training Details
    \end{itemize}
    \item \textbf{S.II. Simulation Experiment Details}
    \begin{itemize}
        \item A. Implementation Details for Depth Map Prediction
        \item B. Robocasa Full Benchmark Results
        \item C. Hyperparameters and Training Details
    \end{itemize}
    \item \textbf{S.III. Additional Experiments}
    \begin{itemize}
        \item A. Additional Analysis on $P_{t}$
        \item B. Comparison to VLAs with implicit 3D awareness
        \item C. Additional Ablations on $\pi_{0.5}$
    \end{itemize}
    \item \textbf{S.IV. Additional Qualitative Results}
    \begin{itemize}
        \item A. 3D Point Track Visualization
        \item B. Real world rollout images
    \end{itemize}
\end{itemize}
Moreover, we provide real world experiment videos in the supplementary materials.
\vspace{0.5cm}

\section{Real World Experiment Details}
\subsection{Real World Setup}
Our robot system overview is illustrated in Figure~\ref{fig:realworld_setup}. We use an OMY robot equipped with a 6-DoF manipulator driven by high precision DYNAMIXEL-Y actuators, supporting payloads up to 3kg, along with a 1-DoF gripper. For perception, we mount an Intel RealSense D405 wrist camera and use Intel RealSense D435 cameras for third-person and top-down views.

\subsection{Real World Experiment Details}
We conduct four real world manipulation tasks to evaluate the model’s spatiotemporal understanding and control. In the main paper Table~VII and appendix Table~\ref{table:realworld_pi_exp}, the columns \textbf{Height, Spatial, Depth, and Tracking} correspond to the four tasks described below.

\boldparagraph{Pick-and-place over an obstacle (Height)}
The instruction is \textit{“Pick up the doll and place it on the right”}. The robot must pick up a doll and place it to the right side of the workspace while avoiding an obstacle in between. We evaluate three obstacle settings: no obstacle, mid-height obstacle, and high obstacle (twice the mid-height). We collect 45 demonstrations and evaluate 10 trials for each setting (30 total).

\boldparagraph{Pick-and-place into a bin (Spatial)}
The instruction is \textit{“Pick up the doll and put it in the white bin”}. We evaluate two conditions. Seen uses in-distribution object placements without distractors. Unseen includes either (i) an additional distractor object at a seen placement or (ii) a target object placed at an unseen location. We collect 45 demonstrations and evaluate 10 trials for Seen and 10 trials for Unseen.

\boldparagraph{Pick the farthest object (Depth)}
The instruction is \textit{“Pick up the hat farthest from the robot”}. Two hats are placed on the table, and the robot must select and grasp the one with the larger distance from the robot base. The target identity is balanced across trials (each hat is the target in half of the trials). For Seen, both hats are placed at in-distribution locations. For Unseen, we report results for two sub-settings indicated in the table as “a/b”: a denotes the case where one of the two hats is at an unseen location, and b denotes the case where both hats are at unseen locations. We collect 48 demonstrations and evaluate 24 trials for Seen and 16 trials for Unseen.

\boldparagraph{Pick a moving object (Tracking)}
The instruction is \textit{“Pick up the Santa hat”}. The target hat is relocated while the robot is approaching, requiring the robot to track the target and execute the grasp at the updated position. Seen uses in-distribution initial and relocated positions. Unseen uses an unseen relocated position (initial position remains in-distribution). OOD additionally relocates the hat twice, inducing stronger distribution shift in the target’s motion. Our dataset includes sequences with a fixed target and sequences with a single relocation. We collect 40 demonstrations and evaluate 12 trials each for Seen, Unseen, and OOD (total 36).

\begin{figure}[h]
  \centering
  \includegraphics[width=0.46\textwidth]{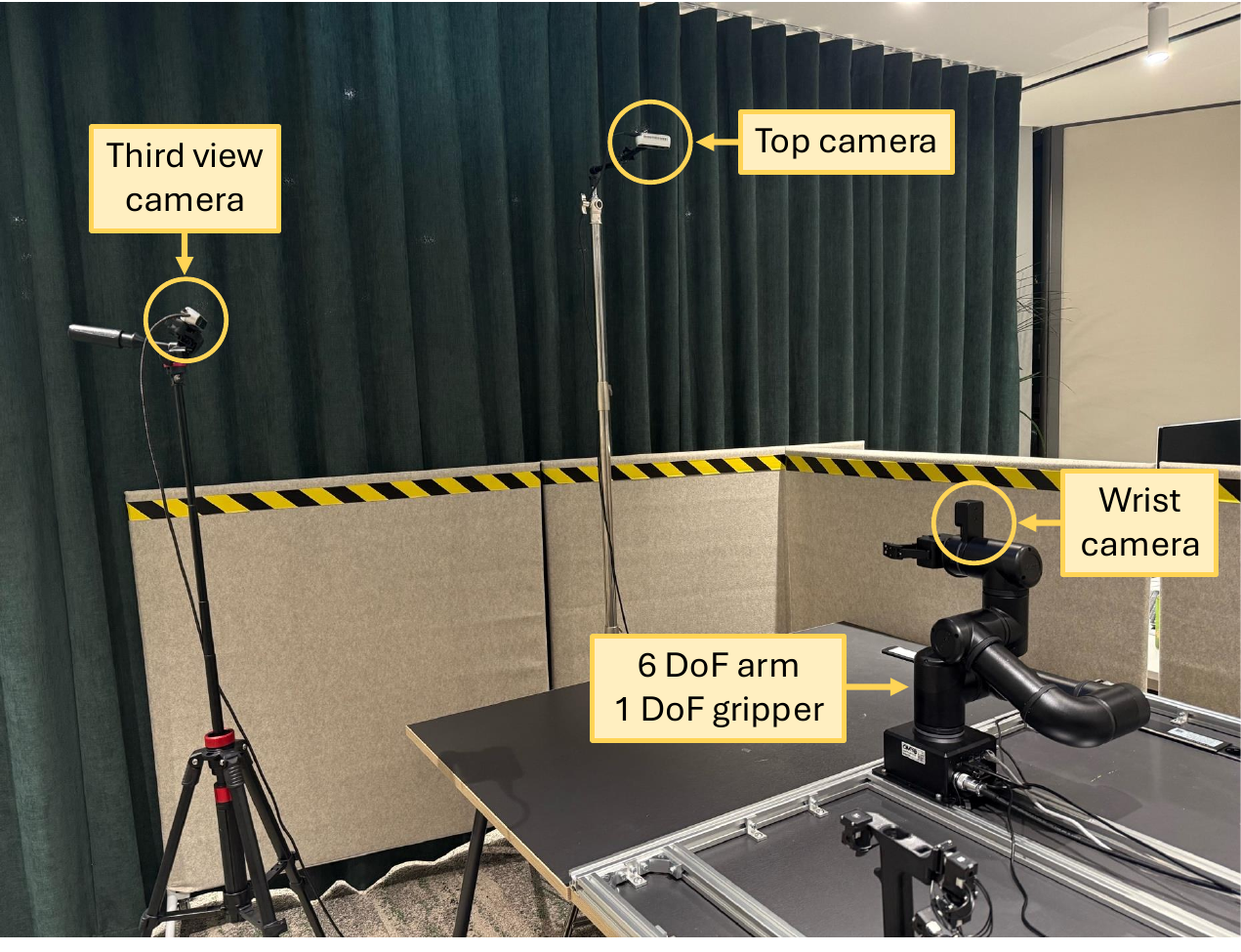}
  \caption{\textbf{Realworld setup}}
  \label{fig:realworld_setup}
\end{figure}

\subsection{Hyperparameters and Training Details}
Table~\ref{tab:hparam_openvla_pri4r_real_compact} and Table~\ref{tab:hparam_pi_pri4r_real_compact} report the hyperparameters used in our real-world training. We follow a shared training protocol across tasks (Height/Spatial/Depth/Tracking) and methods, and modify only the tracking loss weight for Pri4R.
\begin{table}[h]
    \centering
    \caption{\textbf{Hyperparameter details for real-world training.} Unless otherwise specified, OpenVLA-OFT and OpenVLA-OFT+Pri4R share the same settings.}
    \label{tab:hparam_openvla_pri4r_real_compact}
    \resizebox{\columnwidth}{!}{
    \begin{tabular}{lcc}
        \toprule
        & \textbf{OpenVLA-OFT} & \textbf{OpenVLA-OFT + Pri4R} \\
        \midrule
        \multicolumn{3}{l}{\textbf{Shared settings}} \\
        \midrule
        \# GPUs & \multicolumn{2}{c}{4 $\times$ NVIDIA H200} \\
        learning rate (LR) & \multicolumn{2}{c}{$5\mathrm{e}{-4}$} \\
        total batch size & \multicolumn{2}{c}{32 (8 per GPU)} \\
        input images & \multicolumn{2}{c}{3 camera images: third-person, wrist, top-down} \\
        input image size & \multicolumn{2}{c}{224 $\times$ 224} \\
        LoRA rank & \multicolumn{2}{c}{32} \\
        action chunk size & \multicolumn{2}{c}{10} \\
        use proprio (robot state) & \multicolumn{2}{c}{yes} \\
        use FiLM & \multicolumn{2}{c}{no} \\
        \midrule
        \multicolumn{3}{l}{\textbf{Task-specific training steps}} \\
        \midrule
        Height & \multicolumn{2}{c}{60K} \\
        Spatial & \multicolumn{2}{c}{150K} \\
        Depth & \multicolumn{2}{c}{70K} \\
        Tracking & \multicolumn{2}{c}{80K} \\
        \midrule
        \multicolumn{3}{l}{\textbf{Method-specific setting}} \\
        \midrule
        point track loss weight ($\omega_{\mathrm{pt}}$) & -- & 1.0 \\
        \bottomrule
    \end{tabular}}
    \vspace{-2mm}
\end{table}


\begin{table}[h]
    \centering
    \caption{\textbf{Hyperparameter details for training $\pi$ series on real-world tasks.} Unless otherwise specified, $\pi$ and $\pi$+Pri4R share the same settings.}
    \label{tab:hparam_pi_pri4r_real_compact}
    \resizebox{\columnwidth}{!}{%
    \begin{tabular}{lcc}
        \toprule
        & \textbf{$\pi_{0.5}$} & \textbf{$\pi_{0.5}$ + Pri4R} \\
        \midrule
        \multicolumn{3}{l}{\textbf{Shared settings}} \\
        \midrule
        \# GPUs & \multicolumn{2}{c}{4 $\times$ NVIDIA H200} \\
        optimizer & \multicolumn{2}{c}{AdamW} \\
        input images & \multicolumn{2}{c}{3 camera images: third-person, wrist, top-down} \\
        input image size & \multicolumn{2}{c}{224 $\times$ 224} \\
        learning rate scheduler & \multicolumn{2}{c}{Cosine decay} \\
        warmup steps & \multicolumn{2}{c}{10{,}000} \\
        learning rate (LR) & \multicolumn{2}{c}{$5\mathrm{e}{-5}$} \\
        total batch size & \multicolumn{2}{c}{32 (8 per GPU)} \\
        EMA decay & \multicolumn{2}{c}{0.999} \\
        \midrule
        \multicolumn{3}{l}{\textbf{Task-specific training steps}} \\
        \midrule
        Height & \multicolumn{2}{c}{40K} \\
        Spatial & \multicolumn{2}{c}{30K} \\
        Depth & \multicolumn{2}{c}{40K} \\
        Tracking & \multicolumn{2}{c}{30K} \\
        \midrule
        \multicolumn{3}{l}{\textbf{Method-specific setting}} \\
        \midrule
        point track loss weight ($\omega_{pt}$) & -- & 1.0 \\
        \bottomrule
    \end{tabular}%
    }
    \vspace{-2mm}
\end{table}

\section{Simulation Experiment Details}

\subsection{Implementation Details for Depth Map Prediction}
\begin{figure}[h]
    \centering
    \includegraphics[width=0.99\linewidth]{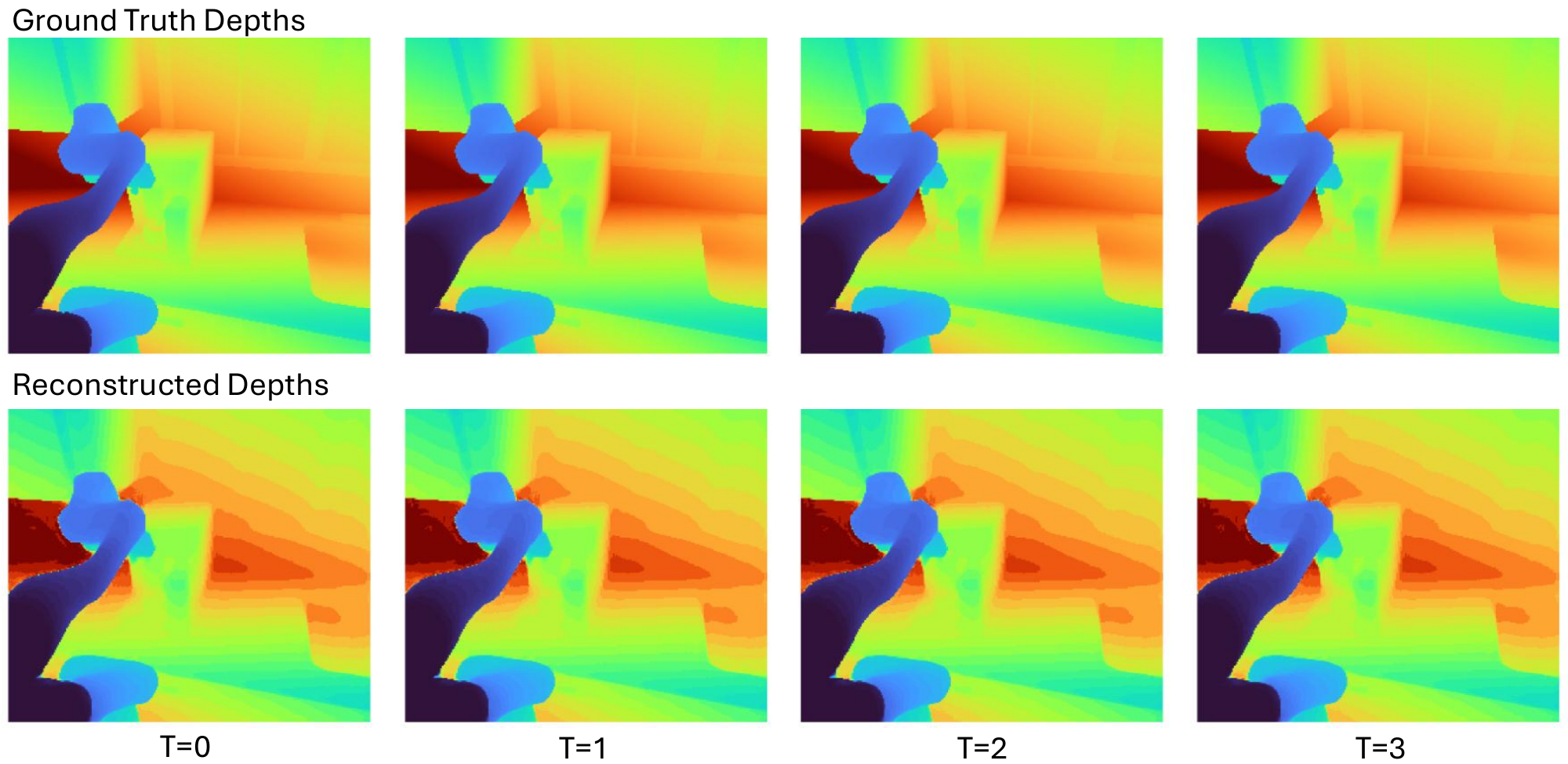}
    \caption{\textbf{Depth reconstruction:} \textbf{(Top)} ground-truth depth maps from the simulator, and \textbf{(Bottom)} depth maps reconstructed by our variational autoencoder (VAE).}
    \vspace{-1em}
    \label{fig:depths}
\end{figure}

In Table~\uppercase\expandafter{\romannumeral 3}, we conduct experiments with alternative supervision targets to analyze the benefit of 3D point tracks. Specifically, to compare against spatially dense depth supervision, we predict future depth maps using a compact latent target. Because raw depth maps are high-resolution, we first train a depth variational autoencoder (VAE) and use its latent codes as supervision. This compresses the representation from $256\times256$ to $32\times32\times4$ (4096 dimensions), which is comparable to the dimensionality of our 3D point-track representation. Figure~\ref{fig:depths} visualizes the depth-VAE reconstructions.

\begin{table}[h]
    \centering
    \caption{\textbf{Simulation training details.} We report settings that differ from real-world training.}
    \label{tab:sim_training_details}
    \resizebox{\columnwidth}{!}{
    \begin{tabular}{lccccc}
        \toprule
        \textbf{Dataset} & \textbf{Method} & \textbf{Input images} & \textbf{\# GPUs} & \textbf{Batch size} & \textbf{\# steps} \\
        \midrule
        \multirow{2}{*}{LIBERO} 
        & OpenVLA-OFT & third-view + wrist-view & 8 & total 64 & 90K \\
        & $\pi_{0}$ / $\pi_{0.5}$          & third-view + wrist-view & 4 & total 128      & 30K \\
        \midrule
        \multirow{2}{*}{RoboCasa}
        & OpenVLA-OFT & left-view + right-view + wrist-view & 8 & total 64 & 120K \\
        & $\pi_{0}$ / $\pi_{0.5}$         & left-view + right-view + wrist-view & 4 & total 128    & 30K \\
        \bottomrule
    \end{tabular}}
    \vspace{-2mm}
\end{table}

\subsection{Robocasa Full Benchmark Results}
We show the full results on Robocasa on Table~\ref{tab:robocasa_full}.

\subsection{Hyperparameters and Training Details}
We follow the same hyperparameter and training protocol as in our real world experiments (Table~\ref{tab:hparam_openvla_pri4r_real_compact} and Table~\ref{tab:hparam_pi_pri4r_real_compact}) unless otherwise specified.
Table~\ref{tab:sim_training_details} summarizes the simulation-specific differences, including camera inputs, compute budget, and training steps for LIBERO and RoboCasa.

\begin{table*}[t]
    \centering
    \caption{\textbf{Robocasa success rates by task.} We report the average success rate (\textbf{Total}) and per-task success rates on RoboCasa for all experiments.}
    \label{tab:robocasa_full}
    \setlength{\tabcolsep}{4.5pt}
    \renewcommand{\arraystretch}{1.05}
    \footnotesize
    \begin{tabular}{l c c c c c c c c}
        \toprule
        \textbf{Method} &
        \textbf{Total} &
        \textbf{PnP} &
        \textbf{Doors} &
        \textbf{Drawers} &
        \textbf{Knobs} &
        \textbf{Lever} &
        \textbf{Press} &
        \textbf{Insert} \\
        \midrule
        OpenVLA-OFT &
        0.331 & 0.218 & 0.457 & 0.590 & 0.080 & 0.360 & 0.560 & 0.270 \\
        \midrule
        OpenVLA-OFT + 2D tracks &
        0.370 & 0.210 & 0.520 & 0.620 & 0.140 & 0.473 & 0.600 & 0.290 \\
        OpenVLA-OFT + Goal point set &
        0.338 & 0.165 & 0.533 & 0.810 & 0.080 & 0.413 & 0.587 & 0.220 \\
        OpenVLA-OFT + Environment points &
        0.352 & 0.190 & 0.487 & 0.470 & 0.170 & 0.540 & 0.493 & 0.280 \\
        OpenVLA-OFT + Robot-only points &
        0.438 & 0.223 & 0.607 & 0.790 & 0.280 & 0.660 & 0.673 & 0.270 \\
        OpenVLA-OFT + \textbf{Ours (Pri4R)} &
        \textbf{0.463} & \textbf{0.230} & \textbf{0.617} & \textbf{0.800} & \textbf{0.250} & \textbf{0.667} & \textbf{0.793} & \textbf{0.340} \\
        \midrule
        $\pi_{0.5}$ &
        0.529 & 0.543 & 0.510 & 0.750 & 0.280 & 0.793 & \textbf{0.600} & 0.040 \\
        \midrule
        $\pi_{0.5}$ + backbone query &
        0.544 & \textbf{0.553} & 0.605 & 0.750 & 0.330 & 0.827 & 0.507 & 0.030 \\
        $\pi_{0.5}$ + query attend action &
        0.548 & 0.535 & 0.605 & 0.840 & \textbf{0.350} & 0.767 & 0.560 & 0.050 \\
        $\pi_{0.5}$ + \textbf{Ours (Pri4R)} &
        \textbf{0.570} & 0.520 & \textbf{0.685} & \textbf{0.890} & 0.330 & \textbf{0.867} & 0.547 & \textbf{0.050} \\
        \bottomrule
    \end{tabular}
\end{table*}

\section{Additional Experiments}

\subsection{Additional Analysis on input $P_t$}
Extending Table~\uppercase\expandafter{\romannumeral 4}, we study a variant that removes the current point set $P_t$ entirely, from both the point track head input and the backbone. This variant still requires no 3D observations at inference time, consistent with Pri4R. However, without $P_t$, the point track head must \emph{generate} future point tracks rather than predict \textit{how a given scene evolves over time}. As shown in Table~\ref{tab:appendix_pt_input_head_ablation}, predicting 3D point tracks without $P_t$ degrades success and can underperform the baseline. This highlights that conditioning the point track head on $P_t$ is crucial for learning world dynamics and, in turn, improving performance.

\begin{table}[h]
    \centering
    \caption{\textbf{Analysis of Point Set $P_t$ input.} We present the average success rates on RoboCasa.}
    \label{tab:appendix_pt_input_head_ablation}
    \setlength{\tabcolsep}{8pt} 
    \begin{tabular}{p{0.66\columnwidth}cc}
        \toprule
        \textbf{Method} & \textbf{SR} $\uparrow$ & \textbf{$\Delta$} \\
        \midrule
        OpenVLA-OFT~\cite{kim2025fine} & 33.1 & -- \\
        \midrule
        + 3D point sequence w/o input pointcloud & 28.7 & -4.4 \\
        \textbf{+ Ours (3D point track)} & \textbf{46.3} & \textbf{+13.2} \\
        \bottomrule
    \end{tabular}
    \vspace{-1em}
\end{table}

\subsection{Comparison with VLAs with implicit 3D awareness}
\begin{table}[t]
    \centering
    \caption{\textbf{Comparison with SpatialForcing.} We compare Pri4R with a recent SOTA method, SpatialForcing.}
    \label{tab:appendix_spatial_forcing}
    \begin{tabular}{l|c|c|c|c|c}
        \toprule
        & Average & Spatial & Object & Goal & Long \\
        \midrule
        OpenVLA-OFT & 92.7 & 92.8 & 98.4 & 96.4 & 83.0 \\
        \midrule
        +SpatialForcing  & 94.2 & \textbf{96.8} & \textbf{99.0} & 96.2 & 84.8 \\
        +\textbf{Pri4R}  & \textbf{95.0} & 93.2 & 96.6 & \textbf{96.8} & \textbf{93.2} \\
        \bottomrule
    \end{tabular}
    \vspace{-3mm}
\end{table}

\boldparagraph{SpatialForcing.}
SpatialForcing~\cite{spatialforcing2025} recently proposed a complementary approach for improving 3D awareness in VLAs by aligning intermediate VLA features with representations from a 3D geometric foundation model, VGGT~\cite{wang2025vggtvisualgeometrygrounded}. Similar to Pri4R, this does not require explicit 3D observations as additional inputs at test time, while also endowing VLAs with additional knowledge.

\begin{table}[h]
    \centering
    \caption{\textbf{Ablations on tracking loss weight for $\pi_{0.5}$.} 
    We present the average success rate (SR) across 24 task suites on RoboCasa, 
    averaged over 50 trials per suite.}
    \label{tab:pi_weight_ablation}

    \resizebox{\columnwidth}{!}{%
    \begin{tabular}{l|c|c|c|c|c|c|c|c}
        \toprule
        weight & Avg & PnP  & Doors & Drawers & Turn & Twist & Press & Insert \\
        \midrule
        0.1  & 54.7 & 54.3 & 62.5 & 90.0 & 28.0 & 73.3 & 55.3 & 3.0 \\
        \rowcolor{OursRow}
        1.0  & 57.0 & 52.0 & 68.5 & 89.0 & 33.0 & 86.7 & 54.7 & 5.0 \\
        10.0 & 50.7 & 47.0 & 54.5 & 77.0 & 35.0 & 82.0 & 47.3 & 5.0 \\
        \bottomrule
    \end{tabular}
    }

    \vspace{-3mm}
\end{table}

\begin{table}[h]
    \centering
    \caption{\textbf{Ablations on number of points for 3D track prediction for $\pi_{0.5}$.} 
    We present the average success rate (SR) across 24 task suites on RoboCasa, 
    averaged over 50 trials per suite.}
    \label{tab:pi_point_number_ablation}

    \resizebox{\columnwidth}{!}{%
    \begin{tabular}{l|c|c|c|c|c|c|c|c}
        \toprule
        number & Avg & PnP  & Doors & Drawers & Turn & Twist & Press & Insert \\
        \midrule
        256  & 48.8 & 49.3 & 55.5 & 77.0 & 28.0 & 73.3 & 41.3 & 1.0 \\
        512  & 53.9 & 48.8 & 60.0 & 87.0 & 27.0 & 81.3 & 59.3 & 7.0 \\
        \rowcolor{OursRow}
        1024 & 57.0 & 52.0 & 68.5 & 89.0 & 33.0 & 86.7 & 54.7 & 5.0 \\
        \bottomrule
    \end{tabular}
    }

    \vspace{-3mm}
\end{table}

Table~\ref{tab:appendix_spatial_forcing} compares SpatialForcing and Pri4R on OpenVLA-OFT. While SpatialForcing yields consistent gains over the baseline, Pri4R achieves higher success rates. We attribute this gap to the nature of the supervision: SpatialForcing primarily injects static 3D structure from an off-the-shelf model, whereas Pri4R leverages temporally dense 4D supervision through 3D point tracks, explicitly encouraging the policy to model interaction dynamics.

\subsection{Additional Ablations}

\boldparagraph{Ablation on $\omega_{\mathrm{pt}}$.}
Table~\ref{tab:pi_weight_ablation} ablates the weight of the auxiliary point track loss, $\omega_{\mathrm{pt}}$. We find that $\omega_{\mathrm{pt}}=1.0$ performs best among the tested values. Notably, because our point track supervision is formulated as displacements (analogous to the action displacement targets), the Pri4R is not highly sensitive to $\omega_{\mathrm{pt}}$ and requires minimal tuning.

\boldparagraph{Ablation on $N_p$.}
Table~\ref{tab:pi_point_number_ablation} ablates the number of tracked points, $N_p$. Although 3D point tracks are intentionally sparse to serve as an efficient supervision signal, using too few points (e.g., 256 or 512) degrades performance compared to our default choice of 1024 points. This suggests that a moderate point density is necessary to sufficiently capture interaction-relevant geometry.

\section{Additional Qualitative Results}
\subsection{3D point track visualization}
\begin{figure*}[t]
    \centering
    \includegraphics[width=0.99\linewidth]{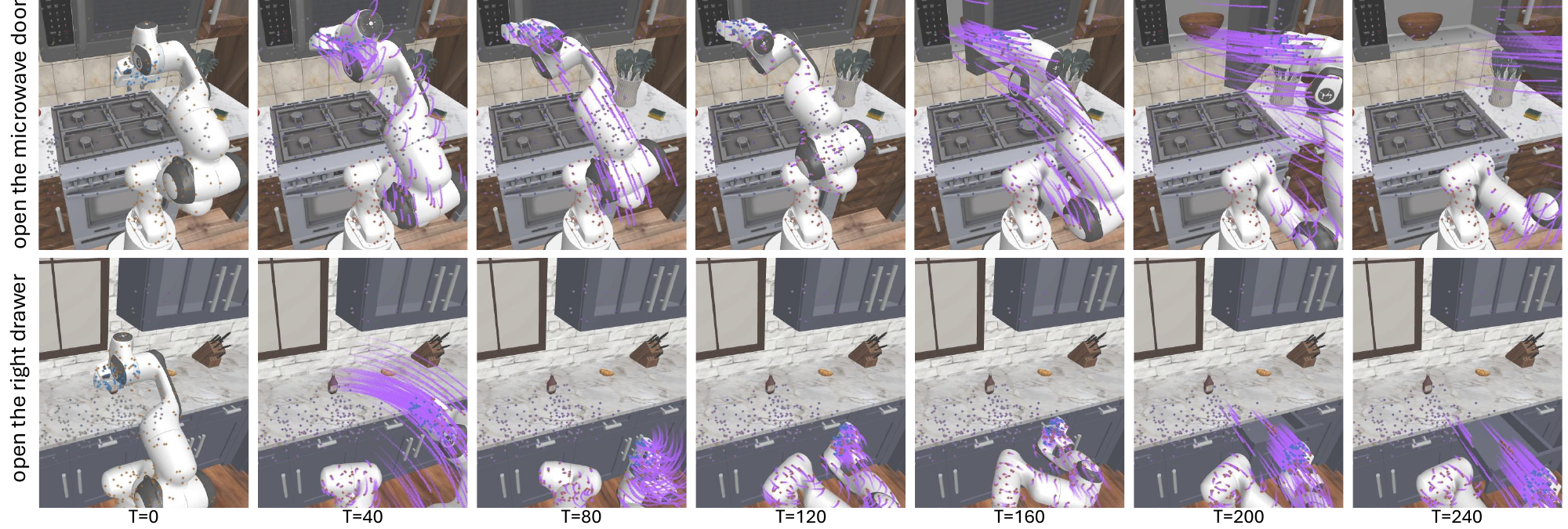}
    \caption{\textbf{Visualization of point track data.} Purple lines denote 3D point tracks. Points are sub-sampled for clarity.}
    \label{fig:point_track_data_vis}
\end{figure*}

In Fig.~\ref{fig:point_track_data_vis}, we visualize the 3D point track data extracted from simulation.

\subsection{Real world rollout images}
\begin{figure*}[t]
    \centering
    \includegraphics[width=0.99\linewidth]{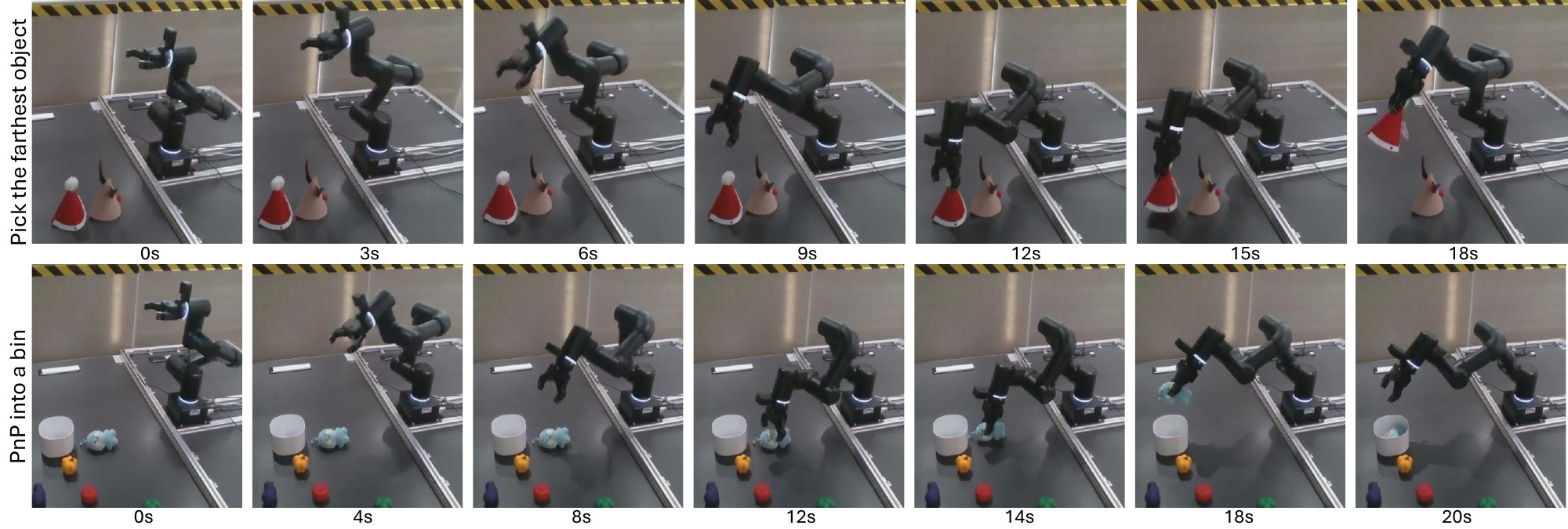}
    \caption{\textbf{Pri4R on real-world tasks.} Additional qualitative results are provided in the supplementary videos.}
    \label{fig:real_world_roll_out}
\end{figure*}

In Fig.~\ref{fig:real_world_roll_out}, we show Pri4R performing real world tasks. Additional qualitative results are provided in the supplementary videos.

\end{document}